\documentclass{article} 
\usepackage{iclr2027_conference,times}

\usepackage{amsmath,amsfonts,bm}

\def\eqref#1{equation~\ref{#1}}

\def\1{\bm{1}}

\DeclareMathAlphabet{\mathsfit}{\encodingdefault}{\sfdefault}{m}{sl}
\SetMathAlphabet{\mathsfit}{bold}{\encodingdefault}{\sfdefault}{bx}{n}

\usepackage{hyperref}
\usepackage{url}
\usepackage{longtable}
\usepackage{amsmath,amssymb}
\usepackage{booktabs,array}
\usepackage{graphicx}
\usepackage{xcolor}
\usepackage{colortbl} 
\usepackage{hyperref}
\usepackage{url}
\providecommand{\method}{LoRA-Norm}
\providecommand{\FPR}{\operatorname{FPR}}
\providecommand{\Energy}{\mathcal{E}}
\providecommand{\diag}{\operatorname{diag}}

\providecommand{\median}{\operatorname{median}}

\usepackage{placeins}
\renewcommand{\eqref}[1]{\textup{(\ref{#1})}}

\usepackage{tabularx, multirow, multicol, graphicx, makecell, enumitem, color, soul, colortbl, booktabs, amssymb}

\title{Learn the Directions, Normalize the Gains: Post-Training Normalization for LoRA}

\author{
Zailong Tian\textsuperscript{1} \quad
Yanzhe Chen\textsuperscript{2} \quad
Zhuoheng Han\textsuperscript{3} \quad
Houfeng Wang\textsuperscript{3,$\dagger$} \quad
Lizi Liao\textsuperscript{1,$\dagger$}
\\[0.25em]
{\small
\textsuperscript{1}Singapore Management University \quad
\textsuperscript{2}National University of Singapore \quad
\textsuperscript{3}Peking University
}
\\[0.15em]
{\small
\texttt{zltian@alumni.pku.edu.cn} \quad
\textsuperscript{$\dagger$}\texttt{wanghf@pku.edu.cn} \quad
\textsuperscript{$\dagger$}\texttt{lzliao@smu.edu.sg}
}
}

\iclrfinalcopy 
\begin{document}

\maketitle
\lhead{}
\begin{abstract}

While Low-Rank Adaptation (LoRA) enables efficient task specialization, its learned updates can compromise capabilities beyond the target task.
We identify \textbf{adaptation imbalance}: a few singular directions dominate the trained update, leaving its performance sensitive to how gains are allocated.
We argue that \textbf{learning where to adapt does not ensure that adaptation gains are well balanced}.
This motivates \textbf{LoRA-Norm}, a post-training normalization method that retains learned directions while rebalancing their gains. 
LoRA-Norm combines spectral rebalancing, a fixed nonlinear transformation of singular values, with nuclear-norm restoration, which preserves the original total spectral mass. It requires no calibration data or additional training and introduces no inference overhead.
Across two backbones and three adaptation tasks, LoRA-Norm improves average specialization and capability retention, outperforming the evaluated post-hoc spectral pruning and gradient-guided editing configurations on both measures. Stronger functional equalization brings no consistent additional gains, revealing that balancing adapter gains and equalizing their responses are distinct objectives.

\end{abstract}







\section{Introduction}
\label{sec:introduction}

\begin{quote}
\itshape
``The concern is not scarcity, but uneven distribution." --- Confucius, \textit{The Analects}
\end{quote}

Low-Rank Adaptation (LoRA) specializes pretrained models by learning compact updates while keeping the original weights frozen \citep{hu2021lora,dettmers2023qlora}. However, freezing pretrained parameters does not preserve pretrained behavior: the added update can improve the target task while compromising capabilities outside the adaptation distribution \citep{shuttleworth2024illusion}. This creates a practical challenge for deployment, where specialization and capability retention both matter~\citep{biderman2024loraless}. An important question is whether a completed adapter can better satisfy these objectives through changes to how its learned directions are used.

We examine this question by separating an adapter's directions from their gains, building on a broader line of parameter-efficient methods that exploit weight decomposition or directional structure~\citep{meng2024pissa,liu2024dora,lingam2024svft}. The decomposition $BA = U\operatorname{diag}(\sigma)V^\top$ expresses the update through learned singular directions and the strengths assigned to them. Although LoRA constrains the update rank, it does not require adaptation strength to be distributed evenly across these directions. Our measurements reveal substantial concentration in trained adapters, with directional output energy dominated by a small subset of the available components. We describe this uneven allocation as \textbf{adaptation imbalance} and investigate its deployment consequences by changing gains while keeping the learned directions fixed.

Our central argument is that \textbf{learning where to adapt does not ensure that adaptation gains are well balanced}. Training jointly determines directions and gains through a source-task objective, but the resulting configuration can remain improvable for specialization and broader capability retention. Prior work establishes the feasibility of post-training refinement through weight interpolation~\citep{wortsman2022wiseft}, spectral pruning \citep{kumaravelu2026para}, and calibration-gradient-guided reweighting \citep{tian2026spectralsurgery}. We investigate a simpler intervention: a fixed normalization rule that revisits the gains of a completed adapter using only its parameters.

\begin{figure}[t]
\centering
\includegraphics[width=0.85\linewidth]{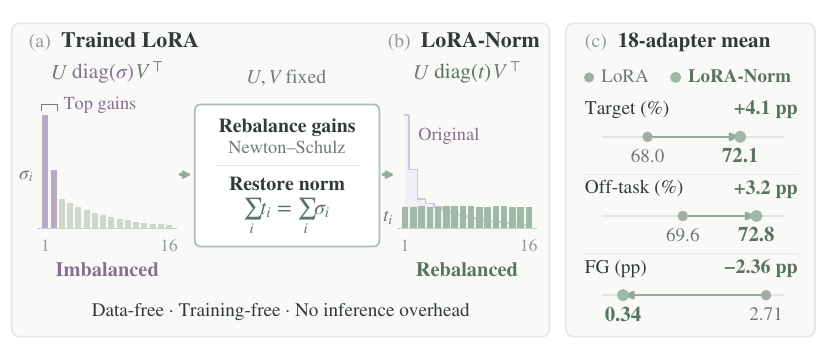}
\caption{
\textbf{Learn the directions, normalize the gains.}
LoRA training learns useful adaptation directions, but the gains assigned
to these directions can become imbalanced.
LoRA-Norm keeps the learned directions fixed and rebalances only their
gains after training, improving specialization while preserving
capabilities beyond the target task.
}
\label{fig:teaser}
\end{figure}

We propose \textbf{LoRA-Norm}, which combines \textit{spectral rebalancing} with \textit{nuclear-norm restoration}, as shown in Fig.~\ref{fig:teaser}. Spectral rebalancing applies a short, fixed sequence of nonlinear transformations to the singular values while preserving their associated directions. Nuclear-norm restoration rescales the transformed gains to preserve their original sum within each module. This changes the allocation of spectral mass without requiring an exactly flat spectrum. The edited update is reconstructed as standard low-rank factors with the original adapter configuration and scaling. LoRA-Norm therefore requires no calibration data or additional training and introduces no inference overhead.

We evaluate 18 trained adapters spanning two backbones and three adaptation tasks in coding, mathematical reasoning, and instruction following. LoRA-Norm improves average target-task performance and capability retention, exceeding the evaluated spectral-pruning and gradient-guided editing configurations on both measures. These results show that the same learned directions can support better deployment performance after gain normalization, without any additional training.

Our analysis also clarifies what this improvement means. Preserving nuclear norm does not preserve update magnitude, and a direction's functional contribution depends on both its gain and the inputs that activate it~\citep{paischer2025eva,yang2024corda}. Controls matched to LoRA-Norm's activation-weighted update energy recover much of its benefit, while stronger functional equalization provides no consistent additional gains. Thus, \textbf{balancing adapter gains and equalizing their responses are distinct objectives}. Directional concentration is useful for characterizing an adapter, but uniformity alone is insufficient to judge whether an edit improves specialization and retention.

In a nutshell, our contributions are summarized as follows:
\begin{itemize}[leftmargin=*]
\item \textbf{Empirical insight.} We characterize adaptation imbalance in trained LoRA adapters and show that their learned directions can support improved specialization and capability retention through post-training gain adjustment.
\item \textbf{Practical method.} We introduce LoRA-Norm, a fixed spectral normalization procedure that preserves learned directions and per-module nuclear mass, requiring neither calibration data nor additional training or inference operations.
\item \textbf{Functional analysis.} We examine gain redistribution alongside update strength through matched-strength controls and activation-aware interventions, showing why stronger response equalization is not a reliable objective for adapter refinement.
\end{itemize}

\section{Related Work}
\label{sec:related_work}

\paragraph{Spectral and direction-aware LoRA training.}
LoRA learns low-rank updates while keeping pretrained weights frozen~\citep{hu2021lora}. PiSSA and MiLoRA initialize trainable factors from principal and minor singular components, respectively~\citep{meng2024pissa,wang2025milora}; DoRA separates weight magnitude and direction~\citep{liu2024dora}; and SCLoRA constrains singular-component growth through spectral clipping~\citep{wi2026sclora}. SVFT learns sparse coefficients, including off-diagonal terms, in a pretrained singular-vector basis~\citep{lingam2024svft}, while LoRA-XS trains a small matrix between frozen SVD-derived factors~\citep{balazy2024loraxs}. These methods change adaptation during training. Spectral analyses also study how update strength relates to intruder directions in merged weights~\citep{shuttleworth2024illusion,xie2026intruderthreshold}. LoRA-Norm instead revisits gains in a completed adapter's learned basis, without further optimization.

\paragraph{Post-training update editing.}
WiSE-FT interpolates pretrained and fine-tuned weights to improve robustness~\citep{wortsman2022wiseft}, corresponding to scaling a LoRA update. Spectral Unforgetting applies hard singular-value thresholding through DG-Hard~\citep{abro2026unforgetting}; SLoRA filters updates by subspace similarity to the base model~\citep{yang2026slora}; and PARA uses a global singular-value threshold for nonuniform rank compression~\citep{kumaravelu2026para}. Safety-oriented methods include S3LoRA, which uses spectral sharpness to guide post-hoc pruning~\citep{ao2025s3lora}, and CSULoRA, which smoothly attenuates update components relative to an estimated safety-aligned subspace~\citep{breneur2026csulora}. Spectral Surgery uses calibration gradients to reweight gains while fixing singular directions, with nuclear-mass restoration~\citep{tian2026spectralsurgery}. LoRA-Norm builds on this fixed-direction formulation, replacing gradient-guided reweighting with a fixed data-free map and studying capability retention and matched-strength controls. LoRA-Norm applies a finite nonlinear transformation with per-module nuclear-norm restoration. Scalar and exact-flattening controls test the roles of update strength and spectral shape.

\paragraph{Activation-aware adaptation and functional analysis.}
Parameter spectra alone do not describe how strongly learned directions respond to inputs. EVA uses activation SVD for LoRA initialization and rank allocation~\citep{paischer2025eva}, while CorDA uses input covariance to guide decomposition toward task learning or knowledge preservation~\citep{yang2024corda}. We use activation statistics retrospectively to characterize completed adapters and construct gain controls. Our functional participation ratio applies the standard participation-ratio form for effective dimensionality~\citep{recanatesi2022dimensionality} to activation-weighted directional energy; it is neither a new mathematical definition nor a direct optimization objective. The main LoRA-Norm edit remains calibration-free.

\section{LoRA-Norm: Post-hoc Spectral Rebalancing}
\label{sec:method}

\begin{figure}[t]
    \centering
    \includegraphics[width=\linewidth]{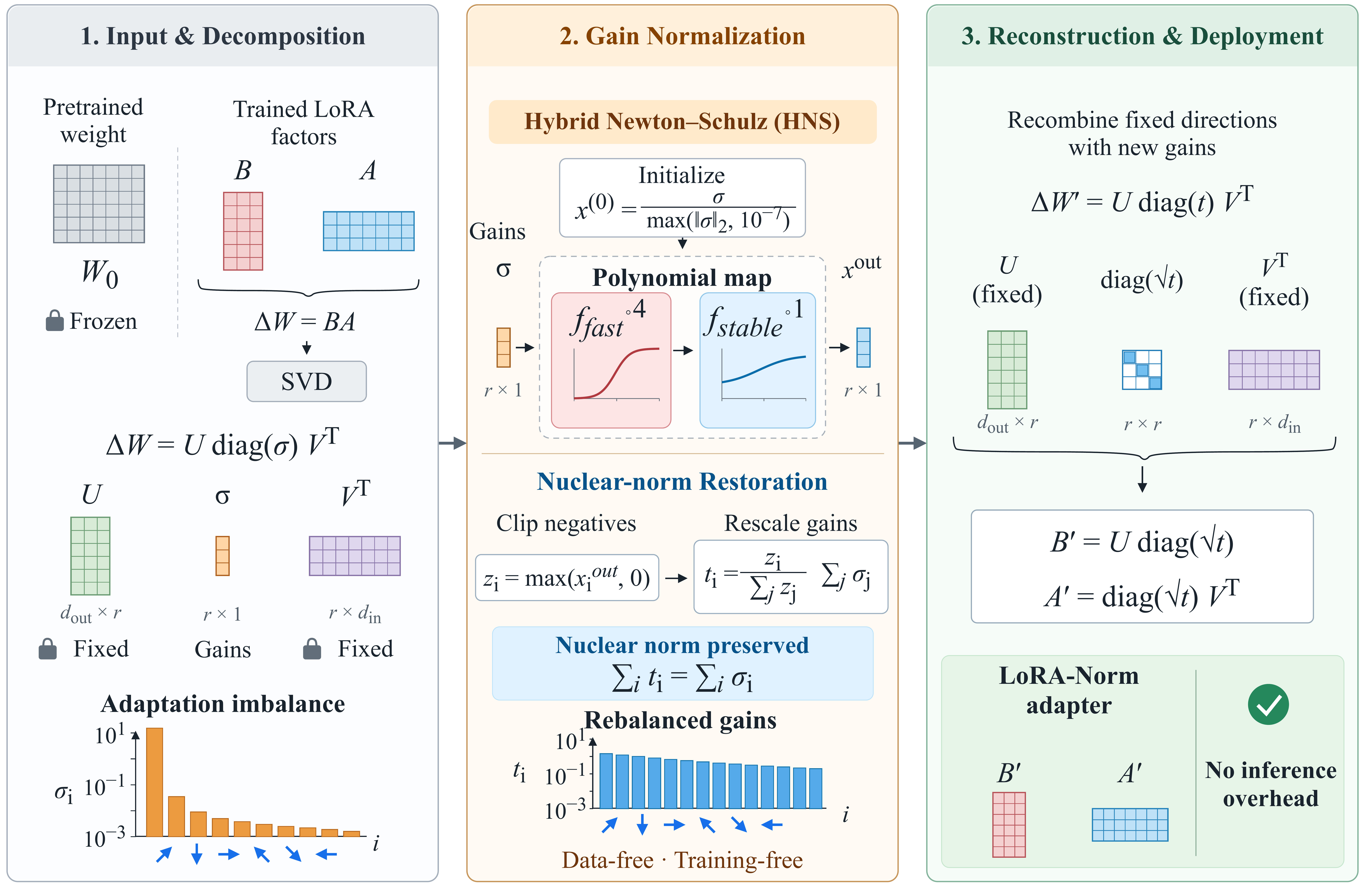}
    \caption{
    \textbf{Overview of LoRA-Norm.}
    A LoRA adapter learns both singular directions and their gains.
    LoRA-Norm keeps the learned directions fixed and applies a post-training spectral
    transformation to rebalance their gains, followed by nuclear-norm restoration.
    The resulting adapter uses the same learned directions with rebalanced gains,
    requiring no additional training or calibration data.
    }
    \label{fig:main}
\end{figure}

A trained LoRA adapter specifies both a set of update directions and the gains assigned to them. We investigate whether these gains can be improved after training, without relearning the directions. \method{} applies a fixed spectral transformation to an existing adapter, preserving its source singular directions and per-module nuclear norm. The editing operation requires neither training data nor gradients; activation statistics
are used separately for analysis. Figure~\ref{fig:main} provides an overview of the
post-training normalization procedure.
\subsection{Fixed-direction adapter editing}
\label{sec:fixed_direction}
For a frozen weight matrix $W_0\in\mathbb{R}^{d_{\mathrm{out}}\times d_{\mathrm{in}}}$, LoRA parameterizes the adapted weight as $W=W_0+cBA$, where $B\in\mathbb{R}^{d_{\mathrm{out}}\times r}$, $A\in\mathbb{R}^{r\times d_{\mathrm{in}}}$, and $c$ is the original adapter scaling~\citep{hu2021lora}. Following the fixed-direction editing formulation of Spectral Surgery~\citep{tian2026spectralsurgery}, we decompose the trained update:
\begin{equation}
D=BA=U\diag(\sigma)V^\top,
\qquad
D'=U\diag(t)V^\top.
\label{eq:fixed_direction}
\end{equation}
Only the nonnegative gains change from $\sigma$ to $t$. We keep their pairing with the source columns of $U,V$ fixed, even when the transformation changes their relative ordering. The base weights, adapter scaling, and stored rank budget remain unchanged.

\subsection{Hybrid Newton--Schulz rebalancing}
\label{sec:hns}
We normalize the source spectrum by its Euclidean norm, $x^{(0)}=\sigma/\max(\|\sigma\|_2,10^{-7})$, and apply elementwise quintic updates in two stages:
\begin{equation}
\begin{aligned}
f_{\mathrm{fast}}(x)&=x\bigl(3.4445-4.7750x^2+2.0315x^4\bigr),\\
f_{\mathrm{stable}}(x)&=x\bigl(2-1.5x^2+0.5x^4\bigr),\\
x^{\mathrm{out}}&=f_{\mathrm{stable}}^{\circ K_s}
                 \bigl(f_{\mathrm{fast}}^{\circ K_f}(x^{(0)})\bigr).
\end{aligned}
\label{eq:hns_iteration}
\end{equation}
Here $f^{\circ K}$ denotes $K$ compositions, not a power. We fix $(K_f,K_s)=(4,1)$ for the main comparisons; Section~\ref{sec:hns_sensitivity} examines other iteration counts. The fast stage strongly amplifies small positive inputs, whereas the stable polynomial has $f_{\mathrm{stable}}(1)=1$ and $f'_{\mathrm{stable}}(1)=0$, refining gains near unity. These local properties motivate the two-stage construction; they do not establish a global convergence or performance guarantee. The fast coefficients follow the quintic transform used in the Muon implementation~\citep{jordanmuon}, but our operation edits a completed adapter rather than a training-time optimizer update.

After clipping numerical negative values, we restore the original nuclear budget and reconstruct balanced low-rank factors:
\begin{equation}
\begin{aligned}
z_i&=\max(x_i^{\mathrm{out}},0), &
t_i&=\frac{\sum_j\sigma_j}{\sum_j z_j}\,z_i,\\
B'&=U\diag(\sqrt{t}), &
A'&=\diag(\sqrt{t})V^\top.
\end{aligned}
\label{eq:reconstruction}
\end{equation}
Thus $\|cB'A'\|_*=\|cBA\|_*$ in exact arithmetic. This is a fixed parameter-space budget, not preservation of the Frobenius norm or functional update energy. Finite-step HNS need not produce an exactly flat spectrum. We obtain the decomposition through reduced QR factorizations and an $r\times r$ core SVD, without materializing the dense update; Appendix~\ref{app:implementation} gives the implementation and cost.

\subsection{Diagnosing functional concentration}
\label{sec:diagnostic}
A parameter gain does not by itself determine how strongly a direction is used. For a module input $h$ from the frozen base model, define $q_i=\mathbb{E}[(v_i^\top h)^2]$. For source-aligned gains $s$ (either $\sigma$ or $t$), orthogonality of $U$ gives
\begin{equation}
E_i(s)=c^2s_i^2q_i,
\qquad
\Energy(s)=\mathbb{E}\|cU\diag(s)V^\top h\|_2^2
          =c^2\sum_i s_i^2q_i.
\label{eq:functional_energy}
\end{equation}
We apply the standard participation-ratio form for effective dimensionality~\citep{recanatesi2022dimensionality} to these directional energies:
\begin{equation}
\pi_i(s)=\frac{s_i^2q_i}{\sum_j s_j^2q_j},
\qquad
\FPR(s;q)=\frac{1}{\sum_i\pi_i(s)^2}.
\label{eq:fpr}
\end{equation}
For nonzero total energy, $1\leq\FPR\leq r$. Larger values indicate energy spread across more source directions. We aggregate module-level FPR by the median within each adapter.

FPR is a diagnostic, not an optimization objective. In particular, $\FPR(as;q)=\FPR(s;q)$ for every $a>0$: it cannot identify changes in overall update strength. We therefore report energy alongside FPR and retain the same frozen-base input statistics across edits. Appendix~\ref{app:diagnostics} also considers the full output second moment to account for correlations between projected activation directions.

\section{Experiments}
\label{sec:experiments}


\subsection{Experimental setup}
\label{sec:setup}
\paragraph{Models and adapters.}
We study Qwen3-8B~\citep{yang2025qwen3} and Llama-3.1-8B-Instruct~\citep{grattafiori2024llama3}, each adapted on three primary tasks: Magicoder~\citep{wei2024magicoder}, MetaMath~\citep{yu2024metamath}, and Tulu-3 Personas-IF~\citep{lambert2024tulu3}, the 29,980-example synthetic instruction-following subset of Tulu 3 (abbreviated Tulu). For each model--task pair, we evaluate three training runs (labeled 42, 43, and 44), yielding 18 source checkpoints. All adapters have rank 16, LoRA alpha 32, and scaling $c=2$, and target the query, key, value, output, gate, up, and down projections. Comparisons pair edits of the same source checkpoint without retraining. \method{} uses the same HNS $4+1$ configuration on all adapter modules at full editing strength in the main comparisons; the separate sensitivity study varies only the iteration counts.
\paragraph{Evaluation.}
The target benchmarks for Magicoder, MetaMath, and Tulu are HumanEval (164 problems; greedy pass@1)~\citep{chen2021code}, GSM8K (1,319 problems; strict accuracy)~\citep{cobbe2021verifiers}, and IFEval (541 prompts; prompt-level strict accuracy)~\citep{zhou2023ifeval}, respectively. We additionally evaluate Commonsense-8, the equal-weight macro accuracy over ARC-Challenge and ARC-Easy~\citep{clark2018arc}, BoolQ~\citep{clark2019boolq}, HellaSwag~\citep{zellers2019hellaswag}, OpenBookQA~\citep{mihaylov2018openbookqa}, PIQA~\citep{bisk2020piqa}, Social IQa~\citep{sap2019socialiqa}, and WinoGrande~\citep{sakaguchi2020winogrande}, totaling 22,419 examples. Commonsense-8 serves as a retention-only category in the primary 18-checkpoint evaluation. Every adapter is evaluated on all four categories using shared benchmark-specific chat templates, parsers, and scoring code. Decoding is greedy with generation seed 42. Appendix~\ref{app:protocol} provides protocol details and token budgets.

\paragraph{Metrics.}
Target is the score on the adapter's target benchmark. Off averages the three non-target categories: the other two target-benchmark categories and Commonsense-8. For model $M$, let $a_j(M)$ denote its percentage score on category $j$, and let $\mathcal O$ contain the three non-target categories. We define
\begin{equation}
\mathrm{Off}(M)=\frac{1}{3}\sum_{j\in\mathcal O}a_j(M),
\qquad
\mathrm{FG}(M)=\frac{1}{3}\sum_{j\in\mathcal O}
\max\!\left\{a_j(\mathrm{Base})-a_j(M),0\right\}.
\label{eq:retention_metrics}
\end{equation}
The forgetting gap (FG) measures category-level deficits relative to the unadapted base model, in percentage points. Commonsense-8 is aggregated before computing Off and FG.

\paragraph{Activation diagnostics.}
For each model--task pair, we collect activations from the frozen base model on 256 fixed source-task training examples, truncated to 512 tokens, using all non-padding tokens. The three runs share the same input examples. For each source checkpoint, all edited variants reuse the activation statistics projected onto its source singular basis. These data are used only for diagnostics and activation-aware controls; \method{} itself is data-free.

\paragraph{Statistical reporting.}
We report means and sample standard deviations over three training runs per model--task pair. Overall results average all 18 source checkpoints equally. Unless otherwise specified, 95\% paired intervals use 2,000 source-checkpoint bootstrap resamples on fixed test sets, keeping all method variants of each checkpoint paired. Appendix~\ref{app:protocol} documents training configurations and differences; Appendices~\ref{app:full_results} and~\ref{app:statistics} provide run-level results and statistical procedures.

\subsection{Target-task performance and capability retention}
\label{sec:main_results}
Averaged over 18 checkpoints, \method{} raises Target from 68.00 to 72.08 ($+4.09$ percentage points) and Off from 69.55 to 72.79 ($+3.24$), while reducing FG from 2.71 to 0.34. Target improves on 16 checkpoints, ties on one, and decreases on one; Off improves on 15 and decreases on three. FG improves on 14 and ties on four. Averaging over the six model--task pairs separately for each run label, Target gains range from 3.86 to 4.30 points and Off gains from 3.17 to 3.30 points.

\begin{table}[!htbp]
\centering
\caption{\textbf{Target-task performance and capability retention.}
Entries are mean $\pm$ sample SD across three training runs;
Overall averages all 18 checkpoints.
Off and FG use the three non-target categories, including Commonsense-8.}
\label{tab:main}
\begingroup
\fontsize{9}{10.5}\selectfont
\definecolor{mtAccent}{HTML}{1F7A6A}
\definecolor{mtMuted}{HTML}{626A70}
\definecolor{mtBand}{HTML}{E5F3EC}
\setlength{\tabcolsep}{3pt}
\renewcommand{\arraystretch}{1.08}
\setlength{\aboverulesep}{2.3pt}
\setlength{\belowrulesep}{2.3pt}
\newcommand{\mtVal}[2]{#1\hspace{0.13em}{\fontsize{7}{8}\selectfont\color{mtMuted}$\pm$#2}}
\newcommand{\mtBest}[2]{\mtVal{\textbf{#1}}{#2}}
\newcommand{\mtMean}[1]{#1\hspace{0.13em}{\fontsize{7}{8}\selectfont\phantom{$\pm$0.00}}}
\newcommand{\mtOurs}{\textcolor{mtAccent}{\textbf{\method{}}}}
\begin{tabular*}{\linewidth}{@{\extracolsep{\fill}}lrr@{\hspace{11pt}}rr@{\hspace{11pt}}rr@{}}
\toprule[0.7pt]
& \multicolumn{2}{c}{\textbf{Target} (\%) $\uparrow$}
& \multicolumn{2}{c}{\textbf{Off} (\%) $\uparrow$}
& \multicolumn{2}{c}{\textbf{FG} (pp) $\downarrow$} \\
\cmidrule(lr){2-3}\cmidrule(lr){4-5}\cmidrule(lr){6-7}
Task & LoRA & \mtOurs & LoRA & \mtOurs & LoRA & \mtOurs \\
\midrule[0.4pt]
\rowcolor{mtBand}
\multicolumn{7}{l}{\rule{0pt}{11pt}\textcolor{mtAccent}{\textbf{Qwen3-8B}}} \\
Magicoder & \mtVal{64.23}{1.27} & \mtBest{74.59}{0.35} & \mtVal{78.04}{0.45} & \mtBest{81.26}{0.07} & \mtVal{1.67}{0.48} & \mtBest{0.00}{0.00} \\
MetaMath & \mtVal{84.05}{0.09} & \mtBest{87.47}{0.68} & \mtVal{68.16}{4.10} & \mtBest{75.28}{1.23} & \mtVal{4.95}{4.05} & \mtBest{0.00}{0.00} \\
Tulu & \mtVal{66.97}{0.83} & \mtBest{71.41}{0.91} & \mtBest{85.63}{0.46} & \mtVal{84.93}{0.44} & \mtBest{0.00}{0.00} & \mtBest{0.00}{0.00} \\
\addlinespace[3pt]
\rowcolor{mtBand}
\multicolumn{7}{l}{\rule{0pt}{11pt}\textcolor{mtAccent}{\textbf{Llama-3.1-8B-Instruct}}} \\
Magicoder & \mtVal{54.47}{1.41} & \mtBest{54.67}{0.93} & \mtVal{61.20}{1.03} & \mtBest{65.80}{0.92} & \mtVal{4.60}{0.52} & \mtBest{1.29}{0.35} \\
MetaMath & \mtVal{75.54}{1.39} & \mtBest{79.35}{1.18} & \mtVal{58.59}{0.93} & \mtBest{62.71}{0.55} & \mtVal{4.07}{0.68} & \mtBest{0.67}{0.89} \\
Tulu & \mtVal{62.72}{0.56} & \mtBest{65.00}{0.21} & \mtVal{65.70}{1.87} & \mtBest{66.77}{0.80} & \mtVal{0.97}{0.95} & \mtBest{0.11}{0.19} \\
\midrule[0.4pt]
\rule{0pt}{11pt}\textcolor{mtAccent}{\textbf{Overall}} & \mtMean{68.00} & \mtMean{\textbf{72.08}} & \mtMean{69.55} & \mtMean{\textbf{72.79}} & \mtMean{2.71} & \mtMean{\textbf{0.34}} \\
\bottomrule[0.7pt]
\end{tabular*}
\endgroup
\end{table}

Qwen--Magicoder gains 10.37 Target points versus 0.20 for Llama--Magicoder (Table~\ref{tab:main}). Off improves in five of six model--task groups, but falls from 85.63 to 84.93 for Qwen--Tulu. FG remains zero in that group: staying above Base does not imply preserving all positive transfer. Appendix~\ref{app:transitions} reports correctness transitions for \method{} and the energy-matched controls.

\subsection{Comparison with post-hoc editing methods}
\label{sec:posthoc_baselines}
We evaluate PARA~\citep{kumaravelu2026para} and Spectral Surgery~\citep{tian2026spectralsurgery} on the same 18 checkpoints under the current protocol. PARA applies global singular-value thresholding with fixed squared-spectral-mass retention targets $\epsilon\in\{0.90,0.95,0.99\}$ and no norm restoration. Spectral Surgery uses the fixed \texttt{grad\_direction} policy, 128 training examples with calibration seed 42, and nuclear-norm restoration. It edits only \texttt{o\_proj/down\_proj}, whereas \method{} edits all seven module families. The comparison therefore concerns these complete configurations and does not isolate gradient guidance. Appendix~\ref{app:posthoc_baselines} provides implementation details and paired intervals.

\begin{table}[!htbp]
\centering
\caption{\textbf{Post-hoc editing methods on 18 checkpoints.}}
\label{tab:posthoc_baselines}
\begingroup
\begin{minipage}{0.94\linewidth}
\centering
\fontsize{9}{10.5}\selectfont
\definecolor{phAccent}{HTML}{1F7A6A}
\definecolor{phMuted}{HTML}{626A70}
\definecolor{phBand}{HTML}{E5F3EC}
\setlength{\tabcolsep}{5pt}
\renewcommand{\arraystretch}{1.13}
\setlength{\aboverulesep}{2.3pt}
\setlength{\belowrulesep}{2.3pt}
\newcommand{\phSetting}[1]{{\color{phMuted}($\epsilon=#1$)}}
\begin{tabular}{>{\raggedright\arraybackslash}p{\dimexpr\linewidth-289.5pt\relax}>{\raggedleft\arraybackslash}p{82pt}*{3}{>{\raggedleft\arraybackslash}p{52.5pt}}}
\toprule[0.7pt]
\textbf{Method} & \textbf{Params retained} (\%) & \textbf{Target} (\%) $\uparrow$ & \textbf{Off} (\%) $\uparrow$ & \textbf{FG} (pp) $\downarrow$ \\
\midrule[0.4pt]
Base & --- & 66.47 & 69.99 & 0.00 \\
LoRA & 100 & 68.00 & 69.55 & 2.71 \\
\midrule[0.4pt]
PARA \phSetting{.90} & 38.16 & 68.78 & 70.01 & 2.37 \\
PARA \phSetting{.95} & 54.28 & 68.58 & 69.64 & 2.70 \\
PARA \phSetting{.99} & 80.16 & 68.44 & 69.78 & 2.56 \\
Spectral Surgery & --- & 68.76 & 68.28 & 3.72 \\
\addlinespace[2.3pt]
\rowcolor{phBand}
\rule{0pt}{11pt}\textcolor{phAccent}{\textbf{\method{}}} & 100 & \textbf{72.08} & \textbf{72.79} & \textbf{0.34} \\
\bottomrule[0.7pt]
\end{tabular}
\end{minipage}
\endgroup
\end{table}
\FloatBarrier

\method{} has the highest mean Target and Off and the lowest FG among the edited adapters in Table~\ref{tab:posthoc_baselines}. It exceeds PARA at $\epsilon=.90$ by 3.31 Target and 2.78 Off points; PARA additionally compresses the adapter, retaining 38.16--80.16\% of parameters across its settings. The evaluated Spectral Surgery configuration improves Target over LoRA by 0.77 points but reduces Off by 1.27. \method{} exceeds it by 3.32 Target and 4.51 Off points without calibration data. These results establish the aggregate ordering of the tested configurations; the matched-strength controls in Section~\ref{sec:s5_shape_strength} examine how much of the benefit can be reproduced by simpler gain changes.

\subsection{Sensitivity to HNS iteration counts}
\label{sec:hns_sensitivity}

We examine iteration-count sensitivity on six source checkpoints, one per backbone for each of the three primary adaptation tasks, varying $K_f\in\{2,4,8\}$ and $K_s\in\{0,1,2\}$. This archived sweep evaluates target-task performance using its own LoRA and $0+0$ reconstruction references and is analyzed separately from the primary 18-checkpoint evaluation. Appendix~\ref{app:hns_step_grid} details the checkpoint selection, evaluation protocol, and complete results. Averaged over these six checkpoints, all nine nonzero configurations yield gains of $+3.42$--$+3.62$ percentage points over LoRA. The main setting $4+1$ yields $+3.57$ points, within $0.05$ points of the largest observed mean. Increasing either the number of fast or stable steps does not monotonically improve performance. The aggregate gain persists across the tested iteration counts, while individual checkpoint responses remain task-dependent.


\section{Analysis}
\label{sec:s5_analysis}


\subsection{RQ1: What changes under gain normalization?}
\label{sec:s5_concentration}

\begin{figure}[!htbp]
\centering
\includegraphics[width=0.8\linewidth]{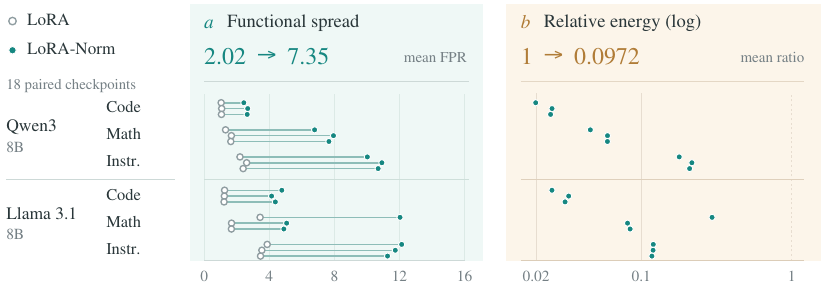}
\caption{\textbf{Gain redistribution and energy reduction.}
(a) Module-median FPR and (b) module-summed energy relative to LoRA under frozen-base inputs.
Each row shows three checkpoints; headers give arithmetic means over all 18.
The dashed line in (b) marks LoRA (=1; log scale).
Code/Math/Instr. denote Magicoder/MetaMath/Tulu.}
\label{fig:s5_concentration}
\end{figure}

Figure~\ref{fig:s5_concentration} summarizes the paired changes across 18 source checkpoints. Using the energy and FPR definitions in Eqs.~\eqref{eq:functional_energy}--\eqref{eq:fpr}, \method{} increases the mean checkpoint-level module-median FPR from 2.02 to 7.35 for rank-16 adapters. The full-moment PR, which also accounts for cross-direction second moments, rises from 1.84 to 5.39 (Appendix~\ref{app:diagnostics}). The mean functional-energy ratio relative to LoRA falls to 0.0972. This ratio sums module energies within each checkpoint before taking the edited-to-original ratio and averaging across checkpoints. The measurements characterize local updates under the fixed frozen-base inputs, rather than the edited network's end-to-end activations.

The coupling between redistribution and strength is already visible in parameter space. For $\Delta W(s)=cU\diag(s)V^\top$, define $S=\sum_i s_i$ and $r_2(s)=S^2/\sum_i s_i^2$. Then
\begin{equation}
\|\Delta W(s)\|_F^2=c^2\frac{S^2}{r_2(s)}.
\label{eq:s5_fro_coupling}
\end{equation}
At fixed nuclear mass, increasing the parameter-spectrum effective rank $r_2$ necessarily lowers the Frobenius norm. Thus, \method{} changes both directional allocation and update strength; nuclear-norm preservation does not make it a shape-only intervention. RQ2 tests the importance of this distinction with matched-strength controls.
\FloatBarrier

\subsection{RQ2: Can matched-strength controls reproduce the gains?}
\label{sec:s5_shape_strength}

\begin{figure}[!htbp]
\centering
\includegraphics[width=0.86\linewidth]{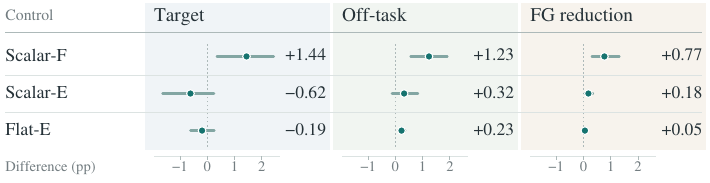}
\caption{\textbf{Matched-strength controls.}
Mean paired differences (pp) across 18 checkpoints with 95\% bootstrap intervals.
Positive favors \method{}: \method{} minus control for Target/Off,
and control minus \method{} for FG.}
\label{fig:s5_controls}
\end{figure}

We construct three controls from each module's actual \method{} spectrum $t$. Scalar-F matches its Frobenius norm, Scalar-E matches its functional energy, and Flat-E uses equal parameter gains at the same functional energy:
\begin{equation}
\begin{aligned}
s_i^{\mathrm{Scalar\text{-}F}}
&=\sigma_i\frac{\|t\|_2}{\|\sigma\|_2},\\
s_i^{\mathrm{Scalar\text{-}E}}
&=\sigma_i\sqrt{\frac{\sum_j t_j^2q_j}{\sum_j\sigma_j^2q_j}},
& s_i^{\mathrm{Flat\text{-}E}}
&=\sqrt{\frac{\sum_jt_j^2q_j}{\sum_jq_j}}.
\end{aligned}
\label{eq:s5_controls}
\end{equation}
The scalar controls preserve relative gains within each module, using a separate multiplier per module. All controls keep the source directions and original LoRA scaling, with no subsequent nuclear-mass restoration. Scalar-F is data-free; Scalar-E and Flat-E require activation statistics. These are diagnostic controls whose matching targets come from \method{}, rather than independently selected deployment strengths. Figure~\ref{fig:s5_controls} shows the paired comparisons; Appendix~\ref{app:matched_strength} gives numerical matching checks and complete results.

\paragraph{Frobenius matching leaves a performance gap.}
\method{} exceeds Scalar-F by 1.44 Target points (95\% paired interval $[0.35,2.44]$) and 1.23 Off points ($[0.56,1.91]$), while reducing FG by 0.77 points. This Frobenius-matching rule does not fully reproduce its average gains. However, Scalar-F's module-level functional energy relative to \method{} has median 2.88: equal Frobenius norms do not imply equal functional strength.

\paragraph{Energy matching reproduces much of the benefit.}
Scalar-E achieves similar aggregate performance (Figure~\ref{fig:s5_controls}). The \method{}-minus-Scalar-E differences are $-0.62$ Target points ($[-1.64,0.25]$) and $+0.32$ Off points ($[-0.13,0.83]$); neither interval excludes zero. \method{} has 0.18 points lower FG ($[0.04,0.35]$). Relative to Flat-E, it gains 0.23 Off points ($[0.10,0.36]$), while the Target difference is $-0.19$ ($[-0.62,0.26]$). These are metric-specific differences, not evidence of uniform superiority or equivalence. Crucially, Scalar-E gains 4.70 Target and 2.92 Off points over LoRA while leaving every module's FPR unchanged. An increase in FPR is therefore not necessary for improvement under these interventions.

\paragraph{Exact parameter flattening is also competitive.}
In a separate evaluation block, Flat-Nuclear preserves each module's nuclear mass and makes its gains equal across the learned singular directions. It improves Target on 16 of 18 checkpoints and trails \method{} by only 0.15 points on average, with paired wins/ties/losses of 8/4/6 against \method{}. This limits claims that the finite-step HNS map is essential. Appendix~\ref{app:parameter_controls} retains that block's own references, Flat-Fro results, and the DG-Hard identity case; its scores are not mixed with the current protocol.

The controls show that functional strength must be considered when interpreting the gains. They do not establish a general shape advantage beyond energy matching. Matching holds under the fixed input statistics of Eq.~\eqref{eq:functional_energy}, so it does not provide a causal decomposition of end-to-end performance. An archived Qwen--MetaMath recheck does not establish HNS superiority over calibration-selected global scaling ($+0.39$ strict points, 95\% interval $[-1.56,2.34]$; Appendix~\ref{app:scalar_calibration}).
\FloatBarrier

\subsection{RQ3: Does stronger functional equalization help?}
\label{sec:s5_equalization}
\label{sec:functional_results}

We test whether pushing directional energies closer to equality improves performance across target-task and off-task evaluations. Functional-HNS protects the second moments using $\bar q_i=\max(q_i,0.1\median(q))$. It weights the input gains by $w_i=(\bar q_i/\median(q))^{\alpha/2}$, applies the same HNS $4+1$ map, divides by $w_i$, and restores the original nuclear mass. We use fixed $\alpha\in\{0.5,1\}$; $\alpha=0$ recovers \method{}. Protected functional flattening instead sets $t_i\propto\bar q_i^{-1/2}$ with the same budget. This flattening exactly equalizes protected directional energies; reported FPR always uses the unfloored $q_i$. Appendix~\ref{app:functional_edits} provides the exact construction of both activation-aware ablations.

\begin{figure}[!htbp]
\centering
\includegraphics[width=0.86\linewidth]{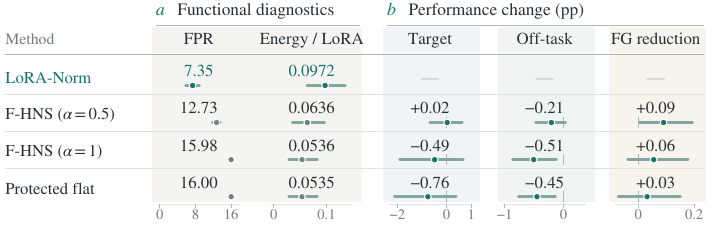}
\caption{\textbf{Stronger functional equalization.}
Diagnostics (a) and paired changes versus \method{} (b, pp).
Positive changes mean higher Target/Off or lower FG.
Dots: means; lines: 95\% checkpoint-bootstrap intervals.
Scales differ; dashes mark the reference.}
\label{fig:s5_equalization}
\end{figure}

\paragraph{Stronger equalization does not consistently help.}
Partial weighting ($\alpha=0.5$) raises mean FPR to 12.73, with only a 0.025-point Target gain and a 0.21-point Off decrease relative to \method{}. Full weighting and protected flattening raise FPR to approximately 16 but lower both mean Target and Off (Figure~\ref{fig:s5_equalization}). Mean FG decreases slightly under all three interventions, so the changes are metric-dependent. Across the 54 checkpoint--intervention comparisons, higher FPR accompanies worse Target performance in 22 cases and worse Off performance in 37. Responses also differ by task: full weighting helps Llama--MetaMath but harms Qwen--Magicoder (Appendix~\ref{app:full_results}). These interventions change energy as well, so they test the usefulness of pursuing stronger equalization without isolating its causal effect on target-task performance and broader capability retention.

\paragraph{FPR adds no stable predictive benefit beyond energy.}
Within-checkpoint centered regressions with leave-one-checkpoint-out evaluation yield the same caution. Among edited variants, adding FPR to log functional-energy ratio changes held-out $R^2$ by 0.0155 on Target and 0.0092 on Off, with both 95\% intervals crossing zero. Within the energy-matched cohort, the changes are $-0.0767$ and $-0.0319$; full-moment PR also fails to give a stable improvement. This is a conditional analysis of relative performance within a checkpoint, using its outcome mean to define the centered target, not label-free prediction of absolute performance. Appendix~\ref{app:fpr_prediction} gives cohorts, fitting procedures, and uncertainty estimates; Figure~\ref{fig:s5_fpr_prediction} visualizes the incremental prediction results. FPR describes how the update is redistributed across the learned singular directions, but these interventions and predictive analyses do not support maximizing it as a performance objective.
\FloatBarrier

\section{Conclusion}
\label{sec:conclusion}
Our results show that the same learned adaptation directions can support better specialization and capability retention after gain adjustment. LoRA-Norm revisits this allocation through fixed spectral rebalancing and per-module nuclear-norm restoration, without calibration data, retraining, or inference overhead. Across 18 adapters spanning two backbones and three tasks, it improves both objectives on average. Energy-matched controls recover much of the benefit, while stronger functional equalization yields no consistent further gains, limiting claims of an independent spectral-shape advantage. Thus, balancing adapter gains and equalizing functional responses are distinct objectives. Broader model and task coverage and further controlled replications would help assess generality.
\FloatBarrier

\subsection*{AI use statement}
The authors conceived the method and designed the experiments. We used OpenAI Codex to assist with code implementation, analysis of experimental outputs, preparation of figures, and manuscript editing. The authors reviewed and executed the AI-assisted code, verified the resulting experimental outputs, and independently assessed the scientific conclusions. The authors take responsibility for the paper's final text, code, results, and scientific claims.

\subsection*{Reproducibility statement}
Section~\ref{sec:setup} describes the experimental setup. Appendix~\ref{app:implementation} provides implementation and mathematical details, and Appendix~\ref{app:protocol} documents evaluation protocols, recorded training configurations, and checkpoint provenance. Appendices~\ref{app:full_results}, \ref{app:matched_strength}, \ref{app:posthoc_baselines}, and~\ref{app:hns_step_grid} report detailed results, control and baseline configurations, and iteration-count sensitivity. Appendix~\ref{app:statistics} specifies the statistical procedures, including paired bootstrap resampling, and describes the machine-readable records underlying the reported analyses. Recorded protocol differences and gaps in archived training information are documented alongside the relevant results.

\bibliography{reference}
\bibliographystyle{iclr2027_conference}

\appendix

\section{Implementation and mathematical details}
\label{app:implementation}
\subsection{Compact decomposition and reconstruction}
Reduced QR factorizations of the two thin factors yield
\begin{equation}
B=Q_BR_B,\qquad A^\top=Q_AR_A,\qquad
R_BR_A^\top=U_r\operatorname{diag}(\sigma)V_r^\top.
\end{equation}
The source singular directions are $U=Q_BU_r$ and $V=Q_AV_r$. For edited nonnegative gains $t$, balanced reconstruction uses $B'=U\operatorname{diag}(\sqrt t)$ and $A'=\operatorname{diag}(\sqrt t)V^\top$, with the original LoRA scaling $c$ retained. Thus the decomposition and balanced reconstruction cost
$O((d_{\mathrm{out}}+d_{\mathrm{in}})r^2+r^3)$ per module, with an additional $O((K_f+K_s)r)$ for the spectral iterations. Working storage is $O((d_{\mathrm{out}}+d_{\mathrm{in}})r+r^2)$, rather than a dense weight-sized update. These are arithmetic complexity bounds, not measured wall-clock speedups.

The compact QR and SVD use float32. Activation-aware spectral editing uses float64 before conversion back to the source spectrum and factor storage types. The reconstructed factors retain the original adapter configuration, including scaling, rank budget, and target modules. Gains are never independently sorted after weighting or iteration, since that would misalign them with the source directions and their activation statistics. Nuclear-budget and reconstruction checks are performed after saving and reloading the edited factors.

\subsection{Norm budgets and exact parameter flattening}
\label{app:flat_definitions}
For gains $s$ and fixed original scaling $c$, the update is $\Delta W=cU\operatorname{diag}(s)V^\top$. Let $S=\sum_i s_i$ and $r_2=S^2/\sum_i s_i^2$. Then
\begin{equation}
\|\Delta W\|_F^2=c^2\frac{S^2}{r_2},\qquad
\sum_i s_i^2\geq\frac{S^2}{r},\qquad
\max_i s_i\geq\frac{S}{r}.
\end{equation}
At fixed nonnegative nuclear mass, the equal-gain endpoint attains both bounds. This algebraic coupling explains why preserving nuclear mass does not preserve Frobenius magnitude; it does not imply optimal task performance. The exact parameter-space controls for source gains $\sigma$ are
\begin{equation}
s_i^{\mathrm{Flat\text{-}Nuclear}}=\frac{\sum_j\sigma_j}{r},\qquad
s_i^{\mathrm{Flat\text{-}Fro}}=\frac{\|\sigma\|_2}{\sqrt r}.
\label{eq:flat_controls}
\end{equation}
The parameter effective rank $r_2$ uses singular-value mass and is not the functional participation ratio defined below. A finite HNS map need not reach the exactly flat endpoint.

\subsection{Functional statistics and their invariances}
\label{app:diagnostics}
Let $C_V=\mathbb{E}[(V^\top h)(V^\top h)^\top]$ be the uncentered second moment of the projected frozen-base inputs. For gains $s$, define
\begin{equation}
G_s=\operatorname{diag}(s)C_V\operatorname{diag}(s),\qquad
\operatorname{FPR}_{\mathrm{full}}(s)=\frac{(\operatorname{tr} G_s)^2}{\operatorname{tr}(G_s^2)}.
\label{eq:full_fpr}
\end{equation}
Writing $q_i=(C_V)_{ii}$ and $e_i=s_i^2q_i$, raw FPR is $(\sum_i e_i)^2/\sum_i e_i^2$, and the module's mean squared output update is $\mathcal E(s)=c^2\sum_i e_i$. The moments are uncentered, not covariances after subtracting the mean. The eigenvalues of $G_s$ give the nonzero eigenvalues of the output update second moment, up to the common factor $c^2$. The raw coordinate-wise FPR uses only the diagonal entries. Since $G_s$ is symmetric,
\begin{equation}
\operatorname{tr}(G_s^2)=\sum_i(G_s)_{ii}^2+\sum_{i\ne j}(G_s)_{ij}^2,
\end{equation}
so $\operatorname{FPR}_{\mathrm{full}}\leq\operatorname{FPR}$ whenever the total energy is nonzero. Equality holds when all off-diagonal entries vanish. Full-moment FPR is invariant to an orthogonal change of coordinates representing the same output second moment; raw FPR depends on the fixed source basis. Neither statistic identifies a common scalar gain.

In particular, for a flat parameter spectrum $s_i=\mu>0$,
\begin{equation}
\operatorname{FPR}(s;q)=\frac{(\sum_iq_i)^2}{\sum_iq_i^2}.
\label{eq:flat_fpr_identity}
\end{equation}
Consequently, Flat-Fro and Flat-Nuclear have identical raw FPR for the same checkpoint, despite their different update magnitudes. Here $q$ depends on both the frozen base activations and the adapter's source directions; it is not a property of the base model alone.

For checkpoint $k$ with modules $m$, we compute
\begin{equation}
R_{E,k}=\frac{\sum_m c_m^2\sum_i t_{mi}^2q_{mi}}
                   {\sum_m c_m^2\sum_i\sigma_{mi}^2q_{mi}},
\qquad R_{\mathrm{RMS},k}=\sqrt{R_{E,k}}.
\end{equation}
Table~\ref{tab:app_functional_summary} reports the mean of $R_{E,k}$, separately from the mean of the per-checkpoint FPR medians. The mean RMS ratios for LoRA, HNS, Functional-HNS ($0.5$), Functional-HNS ($1$), and protected flat are 1.000, 0.288, 0.219, 0.198, and 0.197. Squaring these mean RMS ratios does not give the mean energy ratios.

\subsection{Protected functional edits}
\label{app:functional_edits}
Let $\mathcal H_{4+1}$ denote the HNS gain map used by the main method. For input gains $g$, initialize $x=g/\max(\|g\|_2,10^{-7})$ and apply $x\leftarrow x(a+bx^2+dx^4)$ coordinate-wise: four fast steps use $(a,b,d)=(3.4445,-4.7750,2.0315)$ and one stable step uses $(2,-1.5,0.5)$. Clip numerical negative values and rescale to restore $\sum_i g_i$. Functional-HNS is
\begin{equation}
\begin{aligned}
\bar q_i&=\max(q_i,0.1\operatorname{median}(q)),&
w_i&=\left(\frac{\bar q_i}{\operatorname{median}(q)}\right)^{\alpha/2},\\
\tilde t_i&=\frac{[\mathcal H_{4+1}(\sigma\odot w)]_i}{w_i},&
t_i&=\tilde t_i\frac{\sum_j\sigma_j}{\sum_j\tilde t_j}.
\end{aligned}
\label{eq:app_functional_hns}
\end{equation}
Here $\alpha\in\{0,0.5,1\}$ is a functional weighting exponent, distinct from the LoRA configuration's scaling parameter. After division by $w$, the original parameter-space nuclear sum is restored again. The common intermediate rescaling cancels in the final normalization. For $\alpha=0$, the implementation calls the original helper to retain numerical compatibility.

The protected flattening ablation has the closed form
\begin{equation}
t_i^{\mathrm{PF}}=\frac{\bar q_i^{-1/2}}{\sum_j\bar q_j^{-1/2}}
                  \sum_j\sigma_j,
\qquad \bar q_i=\max(q_i,0.1\operatorname{median}(q)).
\end{equation}
It equalizes $t_i^2\bar q_i$, not necessarily $t_i^2q_i$. Its protected participation ratio is therefore 16, whereas its raw FPR need not be 16. No extra additive epsilon or checkpoint-specific protection ratio is used. Nonfinite inputs, negative gains or second moments, and a nonpositive median trigger an error rather than an activation-free fallback. The median for even rank uses the mean of the two central values.

\section{Evaluation and provenance}
\label{app:protocol}
\subsection{Benchmarks and decoding}
\paragraph{Instruction-following training data.}
Tulu denotes the \texttt{train} split of
\href{https://huggingface.co/datasets/allenai/tulu-3-sft-personas-instruction-following}{\path{allenai/tulu-3-sft-personas-instruction-following}},
the 29,980-example Tulu-3 Personas-IF subset~\citep{lambert2024tulu3}.
It contains synthetic persona-based instructions and assistant responses;
its verifiable constraint types are drawn from the IFEval taxonomy~\citep{zhou2023ifeval}.
All Tulu-labeled results refer to adaptation on this specific subset.

\begin{table}[htbp]
\centering
\small
\caption{Evaluation categories and maximum generated lengths. Commonsense-8 is a macro-average across its eight constituent tasks, not a pooled 22,419-example accuracy.}
\label{tab:protocol}
\begin{tabular}{@{}llrlr@{}}
\toprule
Training task & Target category & Examples & Metric & Tokens \\
\midrule
Magicoder & HumanEval & 164 & Greedy pass@1 & 512 \\
MetaMath & GSM8K & 1,319 & Strict accuracy & 512 \\
Tulu & IFEval & 541 & Prompt strict & 2,048 \\
--- & Commonsense-8 & 22,419 & Macro accuracy & 8 \\
\bottomrule
\end{tabular}
\end{table}

The eight commonsense tasks are ARC-Challenge and ARC-Easy~\citep{clark2018arc}, BoolQ~\citep{clark2019boolq}, HellaSwag~\citep{zellers2019hellaswag}, OpenBookQA~\citep{mihaylov2018openbookqa}, PIQA~\citep{bisk2020piqa}, Social IQa (SIQA)~\citep{sap2019socialiqa}, and WinoGrande~\citep{sakaguchi2020winogrande}. The recorded IFEval protocol evaluates the 541-example artifact named \texttt{ifeval-train.jsonl}; we preserve that artifact rather than substitute another split. This filename is a provenance label, not a claim that evaluation examples were used for training.

All reported generations use temperature 0, top-$p$ 1, generation seed 42, and a maximum context of 4,096 tokens. Qwen is evaluated without thinking mode. HumanEval uses the existing strict-continuation chat protocol; all methods reuse the same benchmark-specific prompt, answer parser, and scoring code. Source run labels do not change generation or calibration seeds. No training-time PEFT alternative is treated as an evaluated baseline in these tables.

\subsection{Activation collection}
For each base--task pair, 256 examples are sampled from the corresponding training SFT data with sampling seed 42 and a 512-token truncation limit. They include the dataset's existing assistant responses and use the base chat template with \texttt{add\_generation\_prompt=False}. All non-padding template, prompt, and assistant tokens contribute to the statistics. We sum projected second moments over tokens and divide by the total valid-token count, rather than averaging examples equally.

The base-only forward pass and projections use bf16, coordinate products use float32, and accumulation uses float64 with TF32 disabled. The three source runs share the same inputs and base forward pass, but use their own source singular bases. Each module caches the full $16\times16$ uncentered moment sum and token counts, and its source basis is verified by a hash. The cache does not contain a full hidden-state covariance for projection onto arbitrary new bases. All edited-adapter functional statistics remain measurements under these frozen-base inputs, not recollected activations of the edited network. No independent retention-distribution calibration or content-level calibration/evaluation overlap result is established by these experiments.

\subsection{Separate evaluation blocks and parameter-space controls}
\label{app:parameter_controls}
\label{sec:spectral_controls}
The primary block contains LoRA, HNS, two Functional-HNS variants, and protected functional flat for 18 sources: 90 records evaluated on four categories, plus eight base-model reference cells, totaling 368 unique cells. During extension to additional runs, a Llama short test exhibited token-level discrepancies. All five methods and the base references for all three Llama runs were therefore re-evaluated together. Qwen references were reused only after token-level agreement checks.

The parameter-space controls in Table~\ref{tab:spectral_controls} come from the earlier joint five-method evaluation of LoRA, Flat-Fro, Flat-Nuclear, DG-Hard, and HNS. That block has its own LoRA/HNS outputs. Comparisons among those controls are internal to the joint block; its scores are not interchangeable with those of the primary block. In particular, the two different HNS aggregates reflect evaluation provenance, not two different choices of HNS iteration counts.

Flat-Fro and Flat-Nuclear use the source singular basis and balanced reconstruction, preserving the Frobenius and nuclear norms, respectively (Eq.~\eqref{eq:flat_controls}). Within this block, \method{} has the highest mean Target, exceeding Flat-Nuclear by 0.151 percentage points. Flat-Nuclear improves 16 of 18 checkpoints over LoRA and wins/ties/loses against \method{} on 8/4/6 checkpoints. Exact flattening therefore recovers much of the observed Target gain without establishing uniform superiority of the finite-step HNS map.

\begin{table}[!htbp]
\centering
\small
\caption{Parameter-space controls in the separate joint evaluation block. All scores and differences in this table use that block's own LoRA/LoRA-Norm references, not those in Table~\ref{tab:main}. Target is averaged across 18 checkpoints; W/T/L denotes paired target wins/ties/losses versus LoRA.}
\label{tab:spectral_controls}
\begin{tabular}{@{}lrrc@{}}
\toprule
Method & Target $\uparrow$ & $\Delta$ vs. LoRA & W/T/L \\
\midrule
LoRA & 68.040 & 0.000 & 0/18/0 \\
Flat-Fro & 71.731 & 3.691 & 16/0/2 \\
Flat-Nuclear & 72.014 & 3.974 & 16/1/1 \\
DG-Hard & 68.040 & 0.000 & 0/18/0 \\
\method{} & \textbf{72.165} & \textbf{4.125} & 17/1/0 \\
\bottomrule
\end{tabular}
\end{table}

For DG-Hard~\citep{abro2026unforgetting}, we apply the Gavish--Donoho median-based hard threshold~\citep{gavish2014threshold}, where $\omega(\beta)$ denotes the complete median-based coefficient (approximately $2.858$ for square matrices) and $\beta$ is the aspect ratio:
\begin{equation}
\tau=\omega(\beta)\operatorname{median}(\sigma_{\mathrm{full}}),\qquad
\beta=\frac{\min(d_{\mathrm{out}},d_{\mathrm{in}})}{\max(d_{\mathrm{out}},d_{\mathrm{in}})}.
\end{equation}
The full spectrum includes the zeros omitted by the low-rank factorization. Its median is zero, making the threshold zero and retaining all 16 active singular values in every source module. DG-Hard is therefore an identity intervention in this regime, rather than evidence against spectral denoising in general. Identity modules keep their original factors byte-for-byte, avoiding a reconstruction perturbation, and are independently generated and scored. Using the median of only the active 16 values would define a different intervention and is not the baseline reported here.

\subsection{Matched-strength evaluation extension}
The later analysis reuses the final primary references and adds Scalar-E and Flat-E for all 18 checkpoints (144 method--category cells). Scalar-F adds 72 cells for those same sources. Thus, the 368-cell count above applies only to the original five-method primary block; the eight-method extension comprises 584 unique method--category/reference cells when counted together. This count excludes the older parameter-control block, the archived step-grid evaluation (Appendix~\ref{app:hns_step_grid}), and the subsequent PARA/Spectral Surgery extension (Appendix~\ref{app:posthoc_baselines}).

Before reuse, source weights/configurations, HNS weights/metadata, activation caches, and evaluation code provenance were checked. For 296 necessary cached cells, input content and ordering were hashed and compared against the corresponding base-model reference. The later Scalar-F inputs were also verified against the final references. Compatibility short tests for the energy-matched extension covered Qwen runs 42/43/44 and Llama runs 43/44; final Llama run-42 caches underwent integrity checks, rather than being justified by an older pilot short test. Input identity verifies the evaluation examples, not equality of model predictions.

The matched-strength extension uses the recorded final prompt, chat-template, parser, and scoring implementations, with a 4,096-token context, 65,536 maximum batched tokens, and adapter blocks of five. The recorded maximum sequence counts are 2,048 for long-generation evaluations and 4,096 for commonsense. Scalar-F evaluation used at most two B300 GPUs; the analyses of existing results and transition counts did not add model inference.

\subsection{Training runs and numerical checks}
Each backbone--task pair has three training runs, labeled 42, 43, and 44. As in Section~\ref{sec:setup}, we report their mean and sample standard deviation ($\mathrm{ddof}=1$), keeping all edited variants of a source checkpoint paired. The standard deviation describes variation across the available runs; configuration differences and incomplete records for the archived run-42 sources are summarized below. Runs 43 and 44 use the same settings within each backbone--task pair except for their training seeds and output locations.

The numerical audit covers all 368 primary cells and 21,420 module-level functional records, with no discrepancy in the recomputed checkpoint medians. The $\alpha=0$ compatibility check has maximum spectral error $4.02\times10^{-8}$, and the reported maximum combined post-save numerical error is $1.80\times10^{-6}$. These checks verify the numerical consistency of the saved artifacts and summaries.

\subsection{Recorded training configurations}
Table~\ref{tab:app_training} reports the recorded settings for runs 43 and 44. Shared settings are rank 16, LoRA alpha 32, dropout 0.05, all-linear adaptation (q/k/v/o and gate/up/down projections), bf16, gradient checkpointing, weight decay 0, gradient clipping 1, Adam $\beta_1=0.9$, and padding to a multiple of eight. Data-subset sampling uses seed 42; both the training seed and Trainer data seed are set to the run label.

Saved training arguments record zero warmup steps in all 12 additional runs; Table~\ref{tab:app_training} reports this executed setting. The Llama math/instruction schedules use minimum learning-rate ratio 0.01. Magicoder's 50,000-example count is before truncation filtering; the retained counts are 49,936 for Qwen and 49,941 for Llama.

\begingroup
\footnotesize
\setlength{\tabcolsep}{4pt}
\begin{longtable}{@{}llrrlrrl@{}}
\caption{Actual settings for each of runs 43 and 44. Batch lists micro-batch times gradient accumulation; each training run uses one GPU. CosMin denotes cosine with a minimum learning rate. Warmup is zero in every row.}\label{tab:app_training}\\
\toprule
Base & Task & Epochs & Length & Batch & LR & $\beta_2$ & Schedule \\
\midrule
\endfirsthead
\multicolumn{8}{l}{\small\itshape Table \thetable\ continued}\\
\toprule
Base & Task & Epochs & Length & Batch & LR & $\beta_2$ & Schedule \\
\midrule
\endhead
\midrule
\multicolumn{8}{r}{\small Continued on next page}\\
\endfoot
\bottomrule
\endlastfoot
Qwen & Code & 1 & 4096 & $16\times2$ & $2\times10^{-5}$ & .999 & Cosine \\
Qwen & Math & 3 & 4096 & $16\times2$ & $10^{-4}$ & .999 & Cosine \\
Qwen & Instr. & 2 & 4096 & $16\times8$ & $4\times10^{-4}$ & .999 & Cosine \\
Llama & Code & 1 & 4096 & $16\times2$ & $2\times10^{-5}$ & .999 & Cosine \\
Llama & Math & 3 & 1024 & $64\times12$ & $10^{-4}$ & .95 & CosMin \\
Llama & Instr. & 2 & 1024 & $64\times2$ & $2\times10^{-4}$ & .95 & CosMin \\
\end{longtable}
\endgroup
For runs 43 and 44, the nominal training sizes are 50,000 for MetaMath and 29,980 for Tulu. Optimizer-update counts are 1,561 for Magicoder, 4,689 for Qwen MetaMath, 198 for Llama MetaMath, and 470 for Tulu. Table~\ref{tab:app_original_training} separately summarizes the available records for run 42.

\begingroup
\small
\setlength{\tabcolsep}{4pt}
\begin{longtable}{@{}p{.19\linewidth}p{.77\linewidth}@{}}
\caption{Recorded settings for the archived run-42 sources. Unverified fields and known differences from runs 43 and 44 are noted explicitly.}\label{tab:app_original_training}\\
\toprule
Source & Recorded settings and differences \\
\midrule
\endfirsthead
\multicolumn{2}{l}{\small\itshape Table \thetable\ continued}\\
\toprule
Source & Recorded settings and differences \\
\midrule
\endhead
\midrule
\multicolumn{2}{r}{\small Continued on next page}\\
\endfoot
\bottomrule
\endlastfoot
Qwen / Code & Model card: 50K examples, 1 epoch, length 4096, global batch 32, LR 2e-5, rank 16, seed 42. Micro-batch and complete optimizer/Trainer settings are unverified. \\
Qwen / Math & Run arguments: 3 epochs, length 4096, global batch 32, LR 1e-4, requested warmup .05, batch $1\times32$. Runs 43/44 use $16\times2$ and zero actual warmup; subset identity across JSON/parquet records is unverified. \\
Qwen / Instr. & Run arguments: 29,980 examples, 2 epochs, length 4096, global batch 128, LR 4e-4, requested warmup .03, batch $1\times128$. Runs 43/44 use $16\times8$, padding to a multiple of eight, and zero actual warmup; subset identity is unverified. \\
Llama / Code & Source documentation: 50K examples, 1 epoch, length 4096, global batch 32, LR 2e-5, rank/alpha 16/32. Training seed and complete Trainer settings are unverified. \\
Llama / Math & Model card: MetaMath 50K, rank 16. Training seed and complete recipe are unverified. \\
Llama / Instr. & Model card: length 1024, global batch 128, rank 16, seed 42. Complete optimizer and schedule settings are unverified. \\
\end{longtable}
\endgroup
The three runs are therefore summarized as available training runs, without attributing all between-run variation to the random seed.

\section{Complete primary results}
\label{app:full_results}
The following tables contain the original five-method block under the final primary evaluation protocol. HNS denotes LoRA-Norm $4+1$, F-HNS denotes Functional-HNS with the indicated exponent, and PF denotes Protected Functional Flat. Qwen and Llama abbreviate Qwen3-8B and Llama-3.1-8B-Instruct; Code, Math, and Instr. denote Magicoder, MetaMath, and Tulu training. All three source runs, including run 42, are retained. Scores are percentages and FG is in percentage points. Means and sample SDs describe the available runs. Values are rounded to three decimals; comparisons use unrounded source values. No best-value highlighting is used as a significance claim.

Table~\ref{tab:functional} gives the aggregate values underlying Figure~\ref{fig:s5_equalization}; Table~\ref{tab:app_functional_summary} additionally reports full-moment PR and RMS ratios. Functional statistics use fixed frozen-base inputs. Performance comparisons and uncertainty estimation retain the checkpoint pairing described in Appendix~\ref{app:statistics}.

\begingroup
\small
\setlength{\tabcolsep}{4pt}
\begin{longtable}{@{}lrrrrr@{}}
\caption{Primary five-method results, averaged over 18 source checkpoints. FPR is the mean checkpoint-level module median, and $R_E$ is the mean functional-energy ratio relative to LoRA, with module energies summed before each checkpoint ratio is formed. Energy ratios are shown to four decimals. Higher Target/Off and lower FG are favorable; FPR and $R_E$ are descriptive diagnostics.}\label{tab:functional}\\
\toprule
Method & Target $\uparrow$ & Off $\uparrow$ & FG $\downarrow$ & FPR & $R_E$ \\
\midrule
\endfirsthead
\multicolumn{6}{l}{\small\itshape Table \thetable\ continued}\\
\toprule
Method & Target $\uparrow$ & Off $\uparrow$ & FG $\downarrow$ & FPR & $R_E$ \\
\midrule
\endhead
\midrule
\multicolumn{6}{r}{\small Continued on next page}\\
\endfoot
\bottomrule
\endlastfoot
LoRA & 67.998 & 69.551 & 2.708 & 2.023 & 1.0000 \\
HNS & 72.083 & 72.792 & 0.344 & 7.345 & 0.0972 \\
F-HNS$_{0.5}$ & 72.108 & 72.586 & 0.254 & 12.728 & 0.0636 \\
F-HNS$_1$ & 71.596 & 72.284 & 0.289 & 15.982 & 0.0536 \\
PF & 71.322 & 72.340 & 0.313 & 16.000 & 0.0535 \\
\end{longtable}
\endgroup

\begingroup
\footnotesize
\setlength{\tabcolsep}{4pt}
\begin{longtable}{@{}llrrrrr@{}}
\caption{Target: three-run mean $\pm$ sample SD (ddof=1).}\label{tab:group_target}\\
\toprule
Base & Task & LoRA & HNS & F-HNS$_{0.5}$ & F-HNS$_1$ & PF \\
\midrule
\endfirsthead
\multicolumn{7}{l}{\small\itshape Table \thetable\ continued}\\
\toprule
Base & Task & LoRA & HNS & F-HNS$_{0.5}$ & F-HNS$_1$ & PF \\
\midrule
\endhead
\midrule
\multicolumn{7}{r}{\small Continued on next page}\\
\endfoot
\bottomrule
\endlastfoot
Qwen & Code & $64.228\pm1.269$ & $74.593\pm0.352$ & $72.358\pm1.535$ & $68.293\pm0.610$ & $68.293\pm0.610$ \\
Qwen & Math & $84.054\pm0.088$ & $87.465\pm0.684$ & $87.617\pm0.789$ & $87.440\pm0.631$ & $87.389\pm0.444$ \\
Qwen & Instr. & $66.975\pm0.834$ & $71.411\pm0.912$ & $72.397\pm0.385$ & $72.828\pm0.640$ & $72.520\pm1.083$ \\
Llama & Code & $54.472\pm1.408$ & $54.675\pm0.931$ & $54.878\pm0.610$ & $55.081\pm0.352$ & $54.472\pm0.352$ \\
Llama & Math & $75.537\pm1.390$ & $79.353\pm1.178$ & $80.642\pm0.684$ & $81.299\pm0.158$ & $80.869\pm0.374$ \\
Llama & Instr. & $62.723\pm0.565$ & $65.003\pm0.213$ & $64.757\pm1.203$ & $64.633\pm0.594$ & $64.387\pm0.930$ \\
\end{longtable}
\endgroup

\begingroup
\footnotesize
\setlength{\tabcolsep}{4pt}
\begin{longtable}{@{}llrrrrr@{}}
\caption{Off: three-run mean $\pm$ sample SD (ddof=1).}\label{tab:group_off}\\
\toprule
Base & Task & LoRA & HNS & F-HNS$_{0.5}$ & F-HNS$_1$ & PF \\
\midrule
\endfirsthead
\multicolumn{7}{l}{\small\itshape Table \thetable\ continued}\\
\toprule
Base & Task & LoRA & HNS & F-HNS$_{0.5}$ & F-HNS$_1$ & PF \\
\midrule
\endhead
\midrule
\multicolumn{7}{r}{\small Continued on next page}\\
\endfoot
\bottomrule
\endlastfoot
Qwen & Code & $78.037\pm0.448$ & $81.261\pm0.071$ & $80.488\pm0.305$ & $80.192\pm0.178$ & $80.163\pm0.178$ \\
Qwen & Math & $68.158\pm4.103$ & $75.277\pm1.235$ & $75.464\pm1.634$ & $76.010\pm1.638$ & $76.002\pm1.389$ \\
Qwen & Instr. & $85.627\pm0.457$ & $84.931\pm0.444$ & $84.855\pm0.514$ & $84.168\pm1.075$ & $84.093\pm0.842$ \\
Llama & Code & $61.198\pm1.030$ & $65.802\pm0.917$ & $65.638\pm1.187$ & $65.643\pm0.373$ & $65.657\pm0.659$ \\
Llama & Math & $58.591\pm0.931$ & $62.710\pm0.554$ & $63.154\pm0.182$ & $62.391\pm0.363$ & $62.301\pm0.044$ \\
Llama & Instr. & $65.696\pm1.870$ & $66.770\pm0.800$ & $65.919\pm0.279$ & $65.301\pm1.155$ & $65.827\pm1.247$ \\
\end{longtable}
\endgroup

\begingroup
\footnotesize
\setlength{\tabcolsep}{4pt}
\begin{longtable}{@{}llrrrrr@{}}
\caption{FG: three-run mean $\pm$ sample SD (ddof=1).}\label{tab:group_fg}\\
\toprule
Base & Task & LoRA & HNS & F-HNS$_{0.5}$ & F-HNS$_1$ & PF \\
\midrule
\endfirsthead
\multicolumn{7}{l}{\small\itshape Table \thetable\ continued}\\
\toprule
Base & Task & LoRA & HNS & F-HNS$_{0.5}$ & F-HNS$_1$ & PF \\
\midrule
\endhead
\midrule
\multicolumn{7}{r}{\small Continued on next page}\\
\endfoot
\bottomrule
\endlastfoot
Qwen & Code & $1.666\pm0.483$ & $0.000\pm0.000$ & $0.000\pm0.000$ & $0.000\pm0.000$ & $0.000\pm0.000$ \\
Qwen & Math & $4.948\pm4.054$ & $0.000\pm0.000$ & $0.000\pm0.000$ & $0.000\pm0.000$ & $0.000\pm0.000$ \\
Qwen & Instr. & $0.000\pm0.000$ & $0.000\pm0.000$ & $0.000\pm0.000$ & $0.000\pm0.000$ & $0.000\pm0.000$ \\
Llama & Code & $4.597\pm0.522$ & $1.289\pm0.351$ & $0.915\pm0.091$ & $0.847\pm0.096$ & $0.901\pm0.174$ \\
Llama & Math & $4.066\pm0.681$ & $0.666\pm0.892$ & $0.458\pm0.740$ & $0.721\pm0.666$ & $0.801\pm0.969$ \\
Llama & Instr. & $0.973\pm0.955$ & $0.109\pm0.190$ & $0.149\pm0.204$ & $0.166\pm0.159$ & $0.173\pm0.160$ \\
\end{longtable}
\endgroup

\begingroup
\footnotesize
\setlength{\tabcolsep}{4pt}
\begin{longtable}{@{}lrlrlrl@{}}
\caption{Paired changes against HNS across 18 sources. Differences are method minus HNS; W/T/L means win/tie/loss, reversing the favorable direction for FG.}\label{tab:paired_outcomes}\\
\toprule
Method & $\Delta$Target & W/T/L & $\Delta$Off & W/T/L & $\Delta$FG & W/T/L \\
\midrule
\endfirsthead
\multicolumn{7}{l}{\small\itshape Table \thetable\ continued}\\
\toprule
Method & $\Delta$Target & W/T/L & $\Delta$Off & W/T/L & $\Delta$FG & W/T/L \\
\midrule
\endhead
\midrule
\multicolumn{7}{r}{\small Continued on next page}\\
\endfoot
\bottomrule
\endlastfoot
F-HNS$_{0.5}$ & 0.025 & 10/0/8 & -0.205 & 7/0/11 & -0.091 & 4/11/3 \\
F-HNS$_1$ & -0.488 & 9/3/6 & -0.507 & 5/0/13 & -0.055 & 4/9/5 \\
PF & -0.762 & 9/1/8 & -0.451 & 5/0/13 & -0.032 & 3/9/6 \\
\end{longtable}
\endgroup

\begingroup
\footnotesize
\setlength{\tabcolsep}{4pt}
\begin{longtable}{@{}llrrrrrr@{}}
\caption{Target: all 18 source checkpoints under the final primary protocol.}\label{tab:seed_target}\\
\toprule
Base & Task & Run & LoRA & HNS & F-HNS$_{0.5}$ & F-HNS$_1$ & PF \\
\midrule
\endfirsthead
\multicolumn{8}{l}{\small\itshape Table \thetable\ continued}\\
\toprule
Base & Task & Run & LoRA & HNS & F-HNS$_{0.5}$ & F-HNS$_1$ & PF \\
\midrule
\endhead
\midrule
\multicolumn{8}{r}{\small Continued on next page}\\
\endfoot
\bottomrule
\endlastfoot
Qwen & Code & 42 & 65.244 & 75.000 & 70.732 & 68.293 & 67.683 \\
Qwen & Code & 43 & 62.805 & 74.390 & 73.780 & 68.902 & 68.902 \\
Qwen & Code & 44 & 64.634 & 74.390 & 72.561 & 67.683 & 68.293 \\
Qwen & Math & 42 & 84.155 & 88.173 & 88.249 & 87.945 & 87.566 \\
Qwen & Math & 43 & 84.003 & 86.808 & 86.732 & 86.732 & 86.884 \\
Qwen & Math & 44 & 84.003 & 87.415 & 87.870 & 87.642 & 87.718 \\
Qwen & Instr. & 42 & 67.837 & 70.795 & 72.089 & 73.198 & 73.752 \\
Qwen & Instr. & 43 & 66.174 & 72.458 & 72.828 & 73.198 & 71.719 \\
Qwen & Instr. & 44 & 66.913 & 70.980 & 72.274 & 72.089 & 72.089 \\
Llama & Code & 42 & 53.659 & 54.878 & 54.268 & 54.878 & 54.878 \\
Llama & Code & 43 & 56.098 & 55.488 & 54.878 & 55.488 & 54.268 \\
Llama & Code & 44 & 53.659 & 53.659 & 55.488 & 54.878 & 54.268 \\
Llama & Math & 42 & 77.104 & 80.667 & 81.350 & 81.425 & 81.122 \\
Llama & Math & 43 & 75.057 & 78.393 & 79.985 & 81.122 & 80.440 \\
Llama & Math & 44 & 74.450 & 78.999 & 80.591 & 81.350 & 81.046 \\
Llama & Instr. & 42 & 63.216 & 64.880 & 65.989 & 64.880 & 65.250 \\
Llama & Instr. & 43 & 62.847 & 65.250 & 63.586 & 63.956 & 64.510 \\
Llama & Instr. & 44 & 62.107 & 64.880 & 64.695 & 65.065 & 63.401 \\
\end{longtable}
\endgroup

\begingroup
\footnotesize
\setlength{\tabcolsep}{4pt}
\begin{longtable}{@{}llrrrrrr@{}}
\caption{FG: all 18 source checkpoints under the final primary protocol.}\label{tab:seed_fg}\\
\toprule
Base & Task & Run & LoRA & HNS & F-HNS$_{0.5}$ & F-HNS$_1$ & PF \\
\midrule
\endfirsthead
\multicolumn{8}{l}{\small\itshape Table \thetable\ continued}\\
\toprule
Base & Task & Run & LoRA & HNS & F-HNS$_{0.5}$ & F-HNS$_1$ & PF \\
\midrule
\endhead
\midrule
\multicolumn{8}{r}{\small Continued on next page}\\
\endfoot
\bottomrule
\endlastfoot
Qwen & Code & 42 & 1.900 & 0.000 & 0.000 & 0.000 & 0.000 \\
Qwen & Code & 43 & 1.986 & 0.000 & 0.000 & 0.000 & 0.000 \\
Qwen & Code & 44 & 1.111 & 0.000 & 0.000 & 0.000 & 0.000 \\
Qwen & Math & 42 & 1.799 & 0.000 & 0.000 & 0.000 & 0.000 \\
Qwen & Math & 43 & 3.523 & 0.000 & 0.000 & 0.000 & 0.000 \\
Qwen & Math & 44 & 9.522 & 0.000 & 0.000 & 0.000 & 0.000 \\
Qwen & Instr. & 42 & 0.000 & 0.000 & 0.000 & 0.000 & 0.000 \\
Qwen & Instr. & 43 & 0.000 & 0.000 & 0.000 & 0.000 & 0.000 \\
Qwen & Instr. & 44 & 0.000 & 0.000 & 0.000 & 0.000 & 0.000 \\
Llama & Code & 42 & 4.978 & 0.910 & 1.020 & 0.876 & 1.102 \\
Llama & Code & 43 & 4.810 & 1.602 & 0.863 & 0.739 & 0.801 \\
Llama & Code & 44 & 4.002 & 1.356 & 0.863 & 0.924 & 0.801 \\
Llama & Math & 42 & 4.705 & 1.690 & 1.312 & 1.467 & 1.910 \\
Llama & Math & 43 & 4.141 & 0.246 & 0.000 & 0.511 & 0.123 \\
Llama & Math & 44 & 3.350 & 0.062 & 0.062 & 0.185 & 0.370 \\
Llama & Instr. & 42 & 1.908 & 0.328 & 0.382 & 0.338 & 0.345 \\
Llama & Instr. & 43 & 0.000 & 0.000 & 0.000 & 0.022 & 0.029 \\
Llama & Instr. & 44 & 1.011 & 0.000 & 0.064 & 0.139 & 0.146 \\
\end{longtable}
\endgroup

\begingroup
\footnotesize
\setlength{\tabcolsep}{4pt}
\begin{longtable}{@{}llrrrrrr@{}}
\caption{Off: all 18 source checkpoints under the final primary protocol.}\label{tab:seed_off}\\
\toprule
Base & Task & Run & LoRA & HNS & F-HNS$_{0.5}$ & F-HNS$_1$ & PF \\
\midrule
\endfirsthead
\multicolumn{8}{l}{\small\itshape Table \thetable\ continued}\\
\toprule
Base & Task & Run & LoRA & HNS & F-HNS$_{0.5}$ & F-HNS$_1$ & PF \\
\midrule
\endhead
\midrule
\multicolumn{8}{r}{\small Continued on next page}\\
\endfoot
\bottomrule
\endlastfoot
Qwen & Code & 42 & 77.798 & 81.333 & 80.407 & 80.018 & 79.967 \\
Qwen & Code & 43 & 77.759 & 81.191 & 80.825 & 80.184 & 80.209 \\
Qwen & Code & 44 & 78.553 & 81.257 & 80.231 & 80.373 & 80.314 \\
Qwen & Math & 42 & 71.293 & 74.920 & 75.249 & 75.925 & 75.703 \\
Qwen & Math & 43 & 69.666 & 76.650 & 77.195 & 77.689 & 77.517 \\
Qwen & Math & 44 & 63.515 & 74.259 & 73.947 & 74.417 & 74.787 \\
Qwen & Instr. & 42 & 86.018 & 85.363 & 85.448 & 84.976 & 84.779 \\
Qwen & Instr. & 43 & 85.125 & 84.954 & 84.573 & 84.579 & 84.346 \\
Qwen & Instr. & 44 & 85.738 & 84.476 & 84.544 & 82.948 & 83.154 \\
Llama & Code & 42 & 60.320 & 64.793 & 64.278 & 65.231 & 64.930 \\
Llama & Code & 43 & 60.943 & 66.028 & 66.164 & 65.739 & 65.830 \\
Llama & Code & 44 & 62.333 & 66.584 & 66.470 & 65.958 & 66.213 \\
Llama & Math & 42 & 57.884 & 63.338 & 63.310 & 62.545 & 62.305 \\
Llama & Math & 43 & 58.245 & 62.293 & 63.198 & 61.977 & 62.343 \\
Llama & Math & 44 & 59.646 & 62.498 & 62.955 & 62.652 & 62.254 \\
Llama & Instr. & 42 & 63.826 & 66.415 & 65.703 & 64.301 & 64.750 \\
Llama & Instr. & 43 & 67.566 & 67.686 & 66.234 & 66.565 & 67.193 \\
Llama & Instr. & 44 & 65.696 & 66.208 & 65.821 & 65.036 & 65.537 \\
\end{longtable}
\endgroup

\begingroup
\small
\setlength{\tabcolsep}{4pt}
\begin{longtable}{@{}lrrrr@{}}
\caption{Functional diagnostics averaged over 18 sources. Each PR is a module median before checkpoint averaging. Energy and RMS are calculated per checkpoint before averaging.}\label{tab:app_functional_summary}\\
\toprule
Method & Raw FPR & Full PR & Energy ratio & RMS ratio \\
\midrule
\endfirsthead
\multicolumn{5}{l}{\small\itshape Table \thetable\ continued}\\
\toprule
Method & Raw FPR & Full PR & Energy ratio & RMS ratio \\
\midrule
\endhead
\midrule
\multicolumn{5}{r}{\small Continued on next page}\\
\endfoot
\bottomrule
\endlastfoot
LoRA & 2.023 & 1.844 & 1.000 & 1.000 \\
HNS & 7.345 & 5.392 & 0.097 & 0.288 \\
F-HNS$_{0.5}$ & 12.728 & 8.831 & 0.064 & 0.219 \\
F-HNS$_1$ & 15.982 & 12.113 & 0.054 & 0.198 \\
PF & 16.000 & 12.130 & 0.053 & 0.197 \\
\end{longtable}
\endgroup

\section{Matched-strength controls}
\label{app:matched_strength}
\subsection{Definitions and numerical verification}
For each module, let $\sigma$ denote the original gains, $t$ the actual HNS $4+1$ gains, and $q_i=\mathbb E[(v_i^\top h)^2]$ the unfloored second moment from the frozen base. Define $E_H=\sum_i t_i^2q_i$. The three controls are
\begin{equation}
\begin{aligned}
s_i^{\mathrm{Scalar\text{-}F}}&=\sigma_i\frac{\|t\|_2}{\|\sigma\|_2},\\
s_i^{\mathrm{Scalar\text{-}E}}&=\sigma_i\sqrt{\frac{E_H}{\sum_j\sigma_j^2q_j}},\\
s_i^{\mathrm{Flat\text{-}E}}&=\sqrt{\frac{E_H}{\sum_jq_j}}.
\end{aligned}
\label{eq:app_matched_controls}
\end{equation}
All retain $U,V$ and original scaling $c$. Actual functional energy includes $c^2$, which cancels in the matching factors. The controls do not subsequently restore nuclear norm, since doing so would change the matching constraint. Scalar-F preserves the source spectral ratios and matches the observed HNS Frobenius norm; it does not target the Frobenius norm of an ideal exact-flat update. Scalar-E preserves those same source ratios while matching functional energy. Flat-E equalizes parameter gains under the functional-energy constraint; it is distinct from the protected functional-flat intervention, which equalizes protected directional energies.

Scalar-F is data-free. Scalar-E and Flat-E use activation statistics, with no protection floor in $q$. The saved factors use balanced reconstruction. Matching is audited per module, not merely after summing all modules. For a matched quantity $Z$, relative error is $|Z_{\mathrm{saved}}/Z_H-1|$. Energy matching holds under the fixed base-input cache; it does not guarantee identical full-network changes under edited-model activations.

\begingroup
\small
\setlength{\tabcolsep}{4pt}
\begin{longtable}{@{}lrr@{}}
\caption{Matching audits on the completed 18-source extension. Analytic energy checks precede storage; saved-factor errors include reconstruction/storage effects.}\label{tab:app_matching_audit}\\
\toprule
Audit & Module rows & Maximum relative error \\
\midrule
\endfirsthead
\multicolumn{3}{l}{\small\itshape Table \thetable\ continued}\\
\toprule
Audit & Module rows & Maximum relative error \\
\midrule
\endhead
\midrule
\multicolumn{3}{r}{\small Continued on next page}\\
\endfoot
\bottomrule
\endlastfoot
Scalar-E / Flat-E: analytic energy & 8568 & $8.882\times10^{-16}$ \\
Scalar-E / Flat-E: saved energy & 8568 & $7.519\times10^{-6}$ \\
Scalar-F: saved Frobenius norm & 4284 & $3.054\times10^{-6}$ \\
\end{longtable}
\endgroup
The per-module Scalar-F/HNS functional-energy ratio has minimum 0.294708, median 2.883298, and maximum 14.773126. Frobenius matching therefore does not imply functional-energy matching. The older name ScalarShrink referred to more than one implementation: one matched actual HNS, while another matched ideal ExactFlatNuclear. Historical weights/results were audited before the missing current-protocol Scalar-F evaluations were run; the old scores are not substituted for the present control.

\subsection{Complete results and paired comparisons}
The four edited methods below use the same final references. Table~\ref{tab:app_matched_overall} also lists original LoRA from that reference block. The current functional-energy results are not pooled with the earlier Flat-Fro/Flat-Nuclear/DG-Hard block. HNS-minus-control differences are computed within each source before averaging. Figure~\ref{fig:s5_controls} visualizes the Target and Off differences and reverses the FG difference to show FG reduction, so positive values favor HNS in all three panels.

\begingroup
\small
\setlength{\tabcolsep}{4pt}
\begin{longtable}{@{}lcrrr@{}}
\caption{Matched-strength controls on 18 source checkpoints. Calibration indicates whether constructing the edit requires activation statistics. Every matching target is derived from the corresponding HNS update. Higher Target/Off and lower FG are favorable.}\label{tab:app_matched_overall}\label{tab:s5_controls}\\
\toprule
Method & Calibration & Target $\uparrow$ & Off $\uparrow$ & FG $\downarrow$ \\
\midrule
\endfirsthead
\multicolumn{5}{l}{\small\itshape Table \thetable\ continued}\\
\toprule
Method & Calibration & Target $\uparrow$ & Off $\uparrow$ & FG $\downarrow$ \\
\midrule
\endhead
\midrule
\multicolumn{5}{r}{\small Continued on next page}\\
\endfoot
\bottomrule
\endlastfoot
LoRA & --- & 67.998 & 69.551 & 2.708 \\
HNS & No & 72.083 & 72.792 & 0.344 \\
Scalar-F & No & 70.643 & 71.558 & 1.111 \\
Scalar-E & Yes & 72.700 & 72.470 & 0.525 \\
Flat-E & Yes & 72.273 & 72.566 & 0.396 \\
\end{longtable}
\endgroup

\begingroup
\small
\setlength{\tabcolsep}{4pt}
\begin{longtable}{@{}llrrrrr@{}}
\caption{Target: complete matched-strength comparison for the 18 source checkpoints.}\label{tab:app_matched_target}\\
\toprule
Base & Task & Run & HNS & Scalar-F & Scalar-E & Flat-E \\
\midrule
\endfirsthead
\multicolumn{7}{l}{\small\itshape Table \thetable\ continued}\\
\toprule
Base & Task & Run & HNS & Scalar-F & Scalar-E & Flat-E \\
\midrule
\endhead
\midrule
\multicolumn{7}{r}{\small Continued on next page}\\
\endfoot
\bottomrule
\endlastfoot
Qwen & Code & 42 & 75.000 & 75.000 & 75.610 & 75.610 \\
Qwen & Code & 43 & 74.390 & 68.902 & 76.829 & 75.000 \\
Qwen & Code & 44 & 74.390 & 69.512 & 77.439 & 75.000 \\
Qwen & Math & 42 & 88.173 & 84.913 & 86.960 & 88.249 \\
Qwen & Math & 43 & 86.808 & 85.064 & 86.126 & 87.339 \\
Qwen & Math & 44 & 87.415 & 85.823 & 85.444 & 87.642 \\
Qwen & Instr. & 42 & 70.795 & 68.762 & 73.013 & 69.871 \\
Qwen & Instr. & 43 & 72.458 & 69.686 & 71.719 & 73.752 \\
Qwen & Instr. & 44 & 70.980 & 68.762 & 71.534 & 71.349 \\
Llama & Code & 42 & 54.878 & 56.098 & 56.098 & 53.049 \\
Llama & Code & 43 & 55.488 & 57.317 & 57.927 & 57.927 \\
Llama & Code & 44 & 53.659 & 57.927 & 59.756 & 54.878 \\
Llama & Math & 42 & 80.667 & 77.710 & 78.620 & 80.970 \\
Llama & Math & 43 & 78.393 & 76.801 & 78.772 & 78.014 \\
Llama & Math & 44 & 78.999 & 76.876 & 78.848 & 78.923 \\
Llama & Instr. & 42 & 64.880 & 64.510 & 66.359 & 64.695 \\
Llama & Instr. & 43 & 65.250 & 64.325 & 63.586 & 63.586 \\
Llama & Instr. & 44 & 64.880 & 63.586 & 63.956 & 65.065 \\
\end{longtable}
\endgroup

\begingroup
\small
\setlength{\tabcolsep}{4pt}
\begin{longtable}{@{}llrrrrr@{}}
\caption{Off: complete matched-strength comparison for the 18 source checkpoints.}\label{tab:app_matched_off}\\
\toprule
Base & Task & Run & HNS & Scalar-F & Scalar-E & Flat-E \\
\midrule
\endfirsthead
\multicolumn{7}{l}{\small\itshape Table \thetable\ continued}\\
\toprule
Base & Task & Run & HNS & Scalar-F & Scalar-E & Flat-E \\
\midrule
\endhead
\midrule
\multicolumn{7}{r}{\small Continued on next page}\\
\endfoot
\bottomrule
\endlastfoot
Qwen & Code & 42 & 81.333 & 79.632 & 81.648 & 81.405 \\
Qwen & Code & 43 & 81.191 & 80.063 & 80.831 & 80.835 \\
Qwen & Code & 44 & 81.257 & 80.051 & 81.613 & 81.292 \\
Qwen & Math & 42 & 74.920 & 73.778 & 73.380 & 74.172 \\
Qwen & Math & 43 & 76.650 & 73.019 & 74.389 & 76.533 \\
Qwen & Math & 44 & 74.259 & 70.264 & 72.868 & 74.477 \\
Qwen & Instr. & 42 & 85.363 & 85.674 & 84.130 & 85.308 \\
Qwen & Instr. & 43 & 84.954 & 85.726 & 85.480 & 84.824 \\
Qwen & Instr. & 44 & 84.476 & 84.755 & 83.916 & 84.131 \\
Llama & Code & 42 & 64.793 & 62.858 & 65.153 & 64.368 \\
Llama & Code & 43 & 66.028 & 64.648 & 66.137 & 66.106 \\
Llama & Code & 44 & 66.584 & 65.618 & 67.054 & 66.500 \\
Llama & Math & 42 & 63.338 & 59.188 & 60.766 & 62.591 \\
Llama & Math & 43 & 62.293 & 60.797 & 62.808 & 61.847 \\
Llama & Math & 44 & 62.498 & 61.488 & 63.051 & 61.940 \\
Llama & Instr. & 42 & 66.415 & 65.895 & 66.463 & 65.990 \\
Llama & Instr. & 43 & 67.686 & 67.611 & 67.291 & 67.823 \\
Llama & Instr. & 44 & 66.208 & 66.978 & 67.475 & 66.049 \\
\end{longtable}
\endgroup

\begingroup
\small
\setlength{\tabcolsep}{4pt}
\begin{longtable}{@{}llrrrrr@{}}
\caption{FG: complete matched-strength comparison for the 18 source checkpoints.}\label{tab:app_matched_fg}\\
\toprule
Base & Task & Run & HNS & Scalar-F & Scalar-E & Flat-E \\
\midrule
\endfirsthead
\multicolumn{7}{l}{\small\itshape Table \thetable\ continued}\\
\toprule
Base & Task & Run & HNS & Scalar-F & Scalar-E & Flat-E \\
\midrule
\endhead
\midrule
\multicolumn{7}{r}{\small Continued on next page}\\
\endfoot
\bottomrule
\endlastfoot
Qwen & Code & 42 & 0.000 & 0.253 & 0.000 & 0.000 \\
Qwen & Code & 43 & 0.000 & 0.101 & 0.000 & 0.000 \\
Qwen & Code & 44 & 0.000 & 0.000 & 0.000 & 0.000 \\
Qwen & Math & 42 & 0.000 & 0.000 & 0.407 & 0.000 \\
Qwen & Math & 43 & 0.000 & 0.493 & 0.000 & 0.000 \\
Qwen & Math & 44 & 0.000 & 3.160 & 1.016 & 0.000 \\
Qwen & Instr. & 42 & 0.000 & 0.000 & 0.000 & 0.000 \\
Qwen & Instr. & 43 & 0.000 & 0.000 & 0.000 & 0.000 \\
Qwen & Instr. & 44 & 0.000 & 0.000 & 0.000 & 0.000 \\
Llama & Code & 42 & 0.910 & 2.441 & 0.903 & 1.057 \\
Llama & Code & 43 & 1.602 & 1.910 & 1.479 & 1.232 \\
Llama & Code & 44 & 1.356 & 1.702 & 1.171 & 1.356 \\
Llama & Math & 42 & 1.690 & 4.010 & 2.636 & 2.234 \\
Llama & Math & 43 & 0.246 & 3.215 & 0.594 & 0.493 \\
Llama & Math & 44 & 0.062 & 2.524 & 0.554 & 0.431 \\
Llama & Instr. & 42 & 0.328 & 0.190 & 0.684 & 0.320 \\
Llama & Instr. & 43 & 0.000 & 0.000 & 0.000 & 0.000 \\
Llama & Instr. & 44 & 0.000 & 0.000 & 0.000 & 0.000 \\
\end{longtable}
\endgroup

\begingroup
\footnotesize
\setlength{\tabcolsep}{4pt}
\begin{longtable}{@{}lllrrr@{}}
\caption{Matched-strength comparisons by backbone and source task: three-run mean $\pm$ sample SD.}\label{tab:app_matched_grouped}\\
\toprule
Base & Task & Method & Target & Off & FG \\
\midrule
\endfirsthead
\multicolumn{6}{l}{\small\itshape Table \thetable\ continued}\\
\toprule
Base & Task & Method & Target & Off & FG \\
\midrule
\endhead
\midrule
\multicolumn{6}{r}{\small Continued on next page}\\
\endfoot
\bottomrule
\endlastfoot
Qwen & Code & HNS & $74.593\pm0.352$ & $81.261\pm0.071$ & $0.000\pm0.000$ \\
Qwen & Code & Scalar-F & $71.138\pm3.358$ & $79.915\pm0.245$ & $0.118\pm0.127$ \\
Qwen & Code & Scalar-E & $76.626\pm0.931$ & $81.364\pm0.462$ & $0.000\pm0.000$ \\
Qwen & Code & Flat-E & $75.203\pm0.352$ & $81.177\pm0.302$ & $0.000\pm0.000$ \\
Qwen & Math & HNS & $87.465\pm0.684$ & $75.277\pm1.235$ & $0.000\pm0.000$ \\
Qwen & Math & Scalar-F & $85.267\pm0.487$ & $72.354\pm1.849$ & $1.218\pm1.700$ \\
Qwen & Math & Scalar-E & $86.176\pm0.759$ & $73.545\pm0.774$ & $0.474\pm0.512$ \\
Qwen & Math & Flat-E & $87.743\pm0.463$ & $75.060\pm1.284$ & $0.000\pm0.000$ \\
Qwen & Instr. & HNS & $71.411\pm0.912$ & $84.931\pm0.444$ & $0.000\pm0.000$ \\
Qwen & Instr. & Scalar-F & $69.070\pm0.534$ & $85.385\pm0.546$ & $0.000\pm0.000$ \\
Qwen & Instr. & Scalar-E & $72.089\pm0.806$ & $84.509\pm0.848$ & $0.000\pm0.000$ \\
Qwen & Instr. & Flat-E & $71.657\pm1.959$ & $84.754\pm0.591$ & $0.000\pm0.000$ \\
Llama & Code & HNS & $54.675\pm0.931$ & $65.802\pm0.917$ & $1.289\pm0.351$ \\
Llama & Code & Scalar-F & $57.114\pm0.931$ & $64.375\pm1.400$ & $2.017\pm0.381$ \\
Llama & Code & Scalar-E & $57.927\pm1.829$ & $66.115\pm0.950$ & $1.184\pm0.288$ \\
Llama & Code & Flat-E & $55.285\pm2.464$ & $65.658\pm1.134$ & $1.215\pm0.150$ \\
Llama & Math & HNS & $79.353\pm1.178$ & $62.710\pm0.554$ & $0.666\pm0.892$ \\
Llama & Math & Scalar-F & $77.129\pm0.505$ & $60.491\pm1.180$ & $3.250\pm0.744$ \\
Llama & Math & Scalar-E & $78.747\pm0.116$ & $62.208\pm1.255$ & $1.262\pm1.191$ \\
Llama & Math & Flat-E & $79.303\pm1.514$ & $62.126\pm0.406$ & $1.053\pm1.023$ \\
Llama & Instr. & HNS & $65.003\pm0.213$ & $66.770\pm0.800$ & $0.109\pm0.190$ \\
Llama & Instr. & Scalar-F & $64.140\pm0.489$ & $66.828\pm0.868$ & $0.063\pm0.110$ \\
Llama & Instr. & Scalar-E & $64.633\pm1.505$ & $67.076\pm0.539$ & $0.228\pm0.395$ \\
Llama & Instr. & Flat-E & $64.449\pm0.770$ & $66.621\pm1.042$ & $0.107\pm0.185$ \\
\end{longtable}
\endgroup

\begingroup
\small
\setlength{\tabcolsep}{4pt}
\begin{longtable}{@{}llrrr@{}}
\caption{All six paired comparisons among the four edits. Differences and 95\% percentile intervals are in percentage points; negative FG differences favor the left method.}\label{tab:app_matched_ci}\\
\toprule
Comparison & Outcome & Mean & CI low & CI high \\
\midrule
\endfirsthead
\multicolumn{5}{l}{\small\itshape Table \thetable\ continued}\\
\toprule
Comparison & Outcome & Mean & CI low & CI high \\
\midrule
\endhead
\midrule
\multicolumn{5}{r}{\small Continued on next page}\\
\endfoot
\bottomrule
\endlastfoot
HNS $-$ Scalar-E & Target & -0.616 & -1.640 & 0.249 \\
HNS $-$ Scalar-E & Off & 0.322 & -0.125 & 0.828 \\
HNS $-$ Scalar-E & FG & -0.181 & -0.346 & -0.037 \\
HNS $-$ Flat-E & Target & -0.190 & -0.616 & 0.259 \\
HNS $-$ Flat-E & Off & 0.225 & 0.104 & 0.356 \\
HNS $-$ Flat-E & FG & -0.052 & -0.138 & 0.025 \\
HNS $-$ Scalar-F & Target & 1.440 & 0.353 & 2.440 \\
HNS $-$ Scalar-F & Off & 1.234 & 0.558 & 1.909 \\
HNS $-$ Scalar-F & FG & -0.767 & -1.317 & -0.307 \\
Scalar-E $-$ Flat-E & Target & 0.426 & -0.470 & 1.352 \\
Scalar-E $-$ Flat-E & Off & -0.097 & -0.569 & 0.375 \\
Scalar-E $-$ Flat-E & FG & 0.129 & 0.024 & 0.266 \\
Scalar-E $-$ Scalar-F & Target & 2.057 & 1.046 & 3.215 \\
Scalar-E $-$ Scalar-F & Off & 0.912 & 0.373 & 1.413 \\
Scalar-E $-$ Scalar-F & FG & -0.586 & -1.028 & -0.203 \\
Flat-E $-$ Scalar-F & Target & 1.630 & 0.518 & 2.690 \\
Flat-E $-$ Scalar-F & Off & 1.008 & 0.350 & 1.655 \\
Flat-E $-$ Scalar-F & FG & -0.715 & -1.221 & -0.297 \\
\end{longtable}
\endgroup

\begingroup
\footnotesize
\setlength{\tabcolsep}{4pt}
\begin{longtable}{@{}lllrrr@{}}
\caption{HNS versus each control within each backbone--task group (three sources per group). Unadjusted descriptive intervals; these small-group results are not used to select favorable configurations.}\label{tab:app_matched_subgroup_ci}\\
\toprule
Group & Comparison & Outcome & Mean & CI low & CI high \\
\midrule
\endfirsthead
\multicolumn{6}{l}{\small\itshape Table \thetable\ continued}\\
\toprule
Group & Comparison & Outcome & Mean & CI low & CI high \\
\midrule
\endhead
\midrule
\multicolumn{6}{r}{\small Continued on next page}\\
\endfoot
\bottomrule
\endlastfoot
Qwen/Code & HNS $-$ Scalar-E & Target & -2.033 & -3.049 & -0.610 \\
Qwen/Code & HNS $-$ Scalar-E & Off & -0.104 & -0.356 & 0.360 \\
Qwen/Code & HNS $-$ Scalar-E & FG & 0.000 & 0.000 & 0.000 \\
Qwen/Code & HNS $-$ Flat-E & Target & -0.610 & -0.610 & -0.610 \\
Qwen/Code & HNS $-$ Flat-E & Off & 0.083 & -0.072 & 0.356 \\
Qwen/Code & HNS $-$ Flat-E & FG & 0.000 & 0.000 & 0.000 \\
Qwen/Code & HNS $-$ Scalar-F & Target & 3.455 & 0.000 & 5.488 \\
Qwen/Code & HNS $-$ Scalar-F & Off & 1.345 & 1.129 & 1.701 \\
Qwen/Code & HNS $-$ Scalar-F & FG & -0.118 & -0.253 & 0.000 \\
Qwen/Math & HNS $-$ Scalar-E & Target & 1.289 & 0.682 & 1.971 \\
Qwen/Math & HNS $-$ Scalar-E & Off & 1.731 & 1.392 & 2.262 \\
Qwen/Math & HNS $-$ Scalar-E & FG & -0.474 & -1.016 & 0.000 \\
Qwen/Math & HNS $-$ Flat-E & Target & -0.278 & -0.531 & -0.076 \\
Qwen/Math & HNS $-$ Flat-E & Off & 0.216 & -0.217 & 0.749 \\
Qwen/Math & HNS $-$ Flat-E & FG & 0.000 & 0.000 & 0.000 \\
Qwen/Math & HNS $-$ Scalar-F & Target & 2.199 & 1.592 & 3.260 \\
Qwen/Math & HNS $-$ Scalar-F & Off & 2.923 & 1.143 & 3.995 \\
Qwen/Math & HNS $-$ Scalar-F & FG & -1.218 & -3.160 & 0.000 \\
Qwen/Instr. & HNS $-$ Scalar-E & Target & -0.678 & -2.218 & 0.739 \\
Qwen/Instr. & HNS $-$ Scalar-E & Off & 0.422 & -0.526 & 1.232 \\
Qwen/Instr. & HNS $-$ Scalar-E & FG & 0.000 & 0.000 & 0.000 \\
Qwen/Instr. & HNS $-$ Flat-E & Target & -0.246 & -1.294 & 0.924 \\
Qwen/Instr. & HNS $-$ Flat-E & Off & 0.177 & 0.055 & 0.345 \\
Qwen/Instr. & HNS $-$ Flat-E & FG & 0.000 & 0.000 & 0.000 \\
Qwen/Instr. & HNS $-$ Scalar-F & Target & 2.341 & 2.033 & 2.773 \\
Qwen/Instr. & HNS $-$ Scalar-F & Off & -0.454 & -0.773 & -0.279 \\
Qwen/Instr. & HNS $-$ Scalar-F & FG & 0.000 & 0.000 & 0.000 \\
Llama/Code & HNS $-$ Scalar-E & Target & -3.252 & -6.098 & -1.220 \\
Llama/Code & HNS $-$ Scalar-E & Off & -0.313 & -0.470 & -0.109 \\
Llama/Code & HNS $-$ Scalar-E & FG & 0.105 & 0.007 & 0.185 \\
Llama/Code & HNS $-$ Flat-E & Target & -0.610 & -2.439 & 1.829 \\
Llama/Code & HNS $-$ Flat-E & Off & 0.144 & -0.078 & 0.311 \\
Llama/Code & HNS $-$ Flat-E & FG & 0.074 & -0.147 & 0.370 \\
Llama/Code & HNS $-$ Scalar-F & Target & -2.439 & -4.268 & -1.220 \\
Llama/Code & HNS $-$ Scalar-F & Off & 1.427 & 0.966 & 1.935 \\
Llama/Code & HNS $-$ Scalar-F & FG & -0.728 & -1.531 & -0.308 \\
Llama/Math & HNS $-$ Scalar-E & Target & 0.607 & -0.379 & 2.047 \\
Llama/Math & HNS $-$ Scalar-E & Off & 0.501 & -0.553 & 2.572 \\
Llama/Math & HNS $-$ Scalar-E & FG & -0.596 & -0.946 & -0.348 \\
Llama/Math & HNS $-$ Flat-E & Target & 0.051 & -0.303 & 0.379 \\
Llama/Math & HNS $-$ Flat-E & Off & 0.583 & 0.446 & 0.747 \\
Llama/Math & HNS $-$ Flat-E & FG & -0.387 & -0.543 & -0.246 \\
Llama/Math & HNS $-$ Scalar-F & Target & 2.224 & 1.592 & 2.957 \\
Llama/Math & HNS $-$ Scalar-F & Off & 2.219 & 1.010 & 4.150 \\
Llama/Math & HNS $-$ Scalar-F & FG & -2.584 & -2.968 & -2.320 \\
Llama/Instr. & HNS $-$ Scalar-E & Target & 0.370 & -1.479 & 1.664 \\
Llama/Instr. & HNS $-$ Scalar-E & Off & -0.307 & -1.267 & 0.395 \\
Llama/Instr. & HNS $-$ Scalar-E & FG & -0.119 & -0.356 & 0.000 \\
Llama/Instr. & HNS $-$ Flat-E & Target & 0.555 & -0.185 & 1.664 \\
Llama/Instr. & HNS $-$ Flat-E & Off & 0.149 & -0.137 & 0.425 \\
Llama/Instr. & HNS $-$ Flat-E & FG & 0.003 & 0.000 & 0.008 \\
Llama/Instr. & HNS $-$ Scalar-F & Target & 0.863 & 0.370 & 1.294 \\
Llama/Instr. & HNS $-$ Scalar-F & Off & -0.058 & -0.769 & 0.520 \\
Llama/Instr. & HNS $-$ Scalar-F & FG & 0.046 & 0.000 & 0.138 \\
\end{longtable}
\endgroup

\section{Statistical and behavioral analyses}
\label{app:statistics}
\subsection{Paired uncertainty estimation}
For performance outcome $J$, the paired estimand for methods $a,b$ is
\begin{equation}
\widehat\Delta_{a,b}=\frac{1}{K}\sum_{k=1}^K[J_k(a)-J_k(b)].
\end{equation}
Each checkpoint's Off score equally averages the three non-target families, with the eight commonsense scores averaged first. FG uses the same family weights after clipping each base-relative deficit at zero. We resample source checkpoints with replacement within the reported group, keeping all method variants paired, and recompute the mean difference. There are 2,000 resamples with NumPy generator seed 20260914; intervals are the empirical 2.5th and 97.5th percentiles. The overall analysis is not additionally stratified by backbone, source task, or run label. Group intervals are unadjusted and descriptive, not multiple-comparison-corrected tests. Fixed test examples and the differences in archived training recipes limit interpretation to the available source collection.

Figure~\ref{fig:s5_equalization} adds descriptive intervals computed from the existing checkpoint records, without new model evaluations. The same 2,000 checkpoint-index resamples (seed 20260914) are shared across all variants and outcomes. Panel (a) reports intervals for the mean checkpoint-level module-median FPR and the mean energy ratio; panel (b) reports intervals for the mean paired performance change relative to HNS. Target and Off use variant minus HNS, whereas FG reduction uses HNS minus variant. Thus the FG-reduction signs in that figure are opposite to the FG-change signs in Table~\ref{tab:paired_outcomes}. These intervals follow the source-checkpoint scheme above and are not additionally stratified. Figure~\ref{fig:s5_controls} instead reproduces the archived matched-strength intervals in Table~\ref{tab:app_matched_ci}; it uses HNS minus control for Target/Off and control minus HNS for FG reduction.

For correctness-transition rates, a separate conditional test-sample bootstrap fixes one source checkpoint. Each atomic benchmark stores the joint five-bit correctness state of Base, LoRA, HNS, Scalar-E, and Flat-E, giving 32 states. Drawing multinomial counts with the observed probabilities and original sample size is equivalent to resampling paired test examples. All edits share each draw, and opportunity-set denominators are recomputed. Undefined differences are omitted and valid-resample coverage is recorded. Independently drawn benchmark strata are combined with the original family weights for that checkpoint's target/off/overall summaries. This procedure also uses 2,000 resamples and seed 20260914. Because checkpoints share evaluation questions, there is no pooled test-sample confidence interval that treats checkpoint--sample events as independent; conditional sample intervals and source-checkpoint intervals describe different uncertainties.

\subsection{Incremental prediction and leave-one-checkpoint-out evaluation}
\label{app:fpr_prediction}
The predefined predictors are raw FPR and full-moment PR, each aggregated by the module median, and log functional-energy ratio $z_E=\log R_E$. The energy ratio sums module energies before forming the checkpoint ratio. Outcomes are performance gains over original LoRA (Target, Off, each family and each commonsense benchmark), and FG reduction. Three fixed cohorts are retained: A contains LoRA, HNS, two Functional-HNS variants, PF, Scalar-E and Flat-E (7 methods, 126 observations); B excludes LoRA (6 methods, 108 observations); C includes only HNS, Scalar-E and Flat-E (3 methods, 54 observations). Every cohort contains the same 18 source checkpoints. Scalar-F was evaluated later and is not added retrospectively.

Let $a$ index methods within source $k$. Before fitting, define
\begin{equation}
\widetilde x_{ka}=x_{ka}-\frac{1}{A_k}\sum_{a'}x_{ka'},\qquad
\widetilde y_{ka}=y_{ka}-\frac{1}{A_k}\sum_{a'}y_{ka'}.
\end{equation}
We fit unregularized least squares without an intercept using Energy only, PR only, or Energy plus PR. Feature standardization uses the standard deviation of the centered columns; near-zero columns are set to zero (threshold $10^{-10}$). Least squares uses relative singular-value cutoff $10^{-10}$. In cohort C, energy is analytically equal across methods and its centered column is exactly zero; storage rounding is not exploited as a predictor. The baseline prediction is zero within-source deviation.

Leave-one-checkpoint-out (LOCO) fitting excludes the held-out source and all its method variants. Predictions are concatenated across the 18 folds to compute
\begin{equation}
R^2_{\mathrm{LOCO}}=1-
\frac{\sum_{k,a}(\widetilde y_{ka}-\widehat y^{(-k)}_{ka})^2}
     {\sum_{k,a}\widetilde y_{ka}^2},\qquad
\Delta R^2=R^2_{E+\mathrm{PR}}-R^2_E.
\label{eq:app_loco}
\end{equation}
The analogous in-sample statistic uses fitted predictions. Negative held-out $R^2$ means worse squared error than predicting zero deviation. Test-group centering uses the held-out group's outcome mean to define its target; this is a conditional within-checkpoint comparison, not label-free prediction of absolute performance on unseen checkpoints. The original implementation standardized using all centered feature rows. An independent training-fold-only standardization audit reproduced 25,920 predictions across the 54 cohort--fold definitions to maximum difference $4.44\times10^{-15}$, with no overlapping source checkpoints. This agreement is expected for these unregularized linear fits and does not remove the outcome-centering qualification.

Intervals for in-sample incremental $R^2$ resample checkpoint groups and refit both models. Intervals for incremental LOCO $R^2$ instead resample checkpoint groups of the already computed paired held-out residuals, recomputing the squared-error difference and denominator; the LOCO models are not refit for each bootstrap draw. These intervals therefore condition on the obtained out-of-fold predictions and do not capture the full variation of retraining the regression procedure. Both use 2,000 draws and the same seed. For within-checkpoint Spearman correlations, values are rounded to eight decimal places to suppress insignificant numerical differences, exact ties receive average ranks, and constant features/outcomes yield undefined correlations. Defined checkpoint correlations are averaged equally and bootstrapped over sources.

\begin{figure}[!htbp]
\centering
\includegraphics[width=\linewidth]{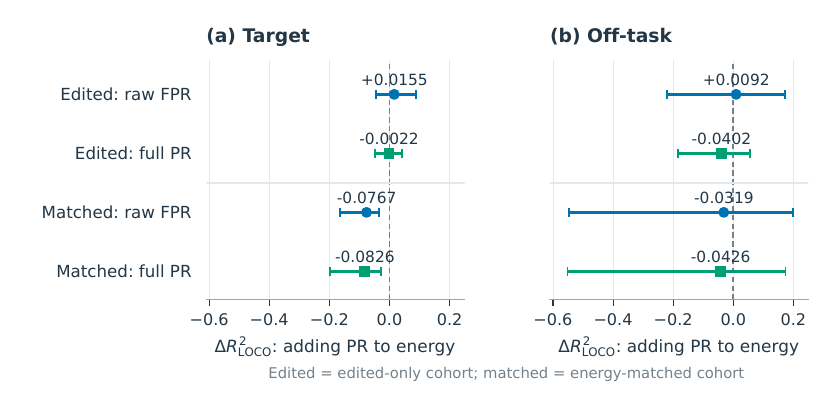}
\caption{\textbf{Incremental predictive value of participation ratios beyond energy.} Points show the change in leave-one-checkpoint-out $R^2$ when raw FPR or full-moment PR is added to log functional-energy ratio, for Target and Off gains. Edited and matched denote cohort B (six edited methods, 108 observations) and cohort C (three energy-matched methods, 54 observations), respectively; both contain 18 source checkpoints. Bars reproduce the reported 95\% bootstrap intervals conditional on the saved held-out predictions. Outcomes are centered within each checkpoint using its outcome mean; this is conditional prediction of relative deviations, not label-free prediction of absolute performance.}
\label{fig:s5_fpr_prediction}
\end{figure}

Figure~\ref{fig:s5_fpr_prediction} summarizes the Target and Off incremental LOCO results for cohorts B and C. In cohort B, the 95\% intervals include zero for both participation-ratio predictors on both outcomes. In cohort C, both Target intervals lie below zero, while both Off intervals include zero. Neither raw FPR nor full-moment PR therefore shows a stable positive increment beyond energy in these comparisons. Tables~\ref{tab:app_fpr_models}--\ref{tab:app_fpr_correlations} retain the complete aggregate-outcome results for all three cohorts.

\FloatBarrier

\begingroup
\footnotesize
\setlength{\tabcolsep}{4pt}
\begin{longtable}{@{}llllrrr@{}}
\caption{Complete Energy/PR/combined fits for the three aggregate outcomes. Full denotes full-moment PR. Repeated Energy rows are the same baseline for each PR comparison; rank is the fitted design rank.}\label{tab:app_fpr_models}\\
\toprule
Cohort & PR & Outcome & Model & $R^2$ & $R^2_{\rm LOCO}$ & Rank \\
\midrule
\endfirsthead
\multicolumn{7}{l}{\small\itshape Table \thetable\ continued}\\
\toprule
Cohort & PR & Outcome & Model & $R^2$ & $R^2_{\rm LOCO}$ & Rank \\
\midrule
\endhead
\midrule
\multicolumn{7}{r}{\small Continued on next page}\\
\endfoot
\bottomrule
\endlastfoot
A & Raw & Target gain & Energy & 0.2120 & 0.1927 & 1 \\
A & Raw & Target gain & FPR & 0.0148 & -0.0206 & 1 \\
A & Raw & Target gain & Energy+FPR & 0.3004 & 0.2127 & 2 \\
A & Full & Target gain & Energy & 0.2120 & 0.1927 & 1 \\
A & Full & Target gain & FPR & 0.0077 & -0.0276 & 1 \\
A & Full & Target gain & Energy+FPR & 0.3235 & 0.2404 & 2 \\
A & Raw & Off gain & Energy & 0.3912 & 0.3532 & 1 \\
A & Raw & Off gain & FPR & 0.0866 & 0.0539 & 1 \\
A & Raw & Off gain & Energy+FPR & 0.4391 & 0.3859 & 2 \\
A & Full & Off gain & Energy & 0.3912 & 0.3532 & 1 \\
A & Full & Off gain & FPR & 0.0647 & 0.0364 & 1 \\
A & Full & Off gain & Energy+FPR & 0.4603 & 0.4091 & 2 \\
A & Raw & FG reduction & Energy & 0.5171 & 0.4730 & 1 \\
A & Raw & FG reduction & FPR & 0.1724 & 0.1531 & 1 \\
A & Raw & FG reduction & Energy+FPR & 0.5371 & 0.4860 & 2 \\
A & Full & FG reduction & Energy & 0.5171 & 0.4730 & 1 \\
A & Full & FG reduction & FPR & 0.1526 & 0.1368 & 1 \\
A & Full & FG reduction & Energy+FPR & 0.5439 & 0.4909 & 2 \\
B & Raw & Target gain & Energy & 0.1867 & 0.0706 & 1 \\
B & Raw & Target gain & FPR & 0.1000 & -0.0049 & 1 \\
B & Raw & Target gain & Energy+FPR & 0.2092 & 0.0861 & 2 \\
B & Full & Target gain & Energy & 0.1867 & 0.0706 & 1 \\
B & Full & Target gain & FPR & 0.1158 & 0.0162 & 1 \\
B & Full & Target gain & Energy+FPR & 0.1973 & 0.0684 & 2 \\
B & Raw & Off gain & Energy & 0.0601 & 0.0241 & 1 \\
B & Raw & Off gain & FPR & 0.0078 & -0.0658 & 1 \\
B & Raw & Off gain & Energy+FPR & 0.1416 & 0.0332 & 2 \\
B & Full & Off gain & Energy & 0.0601 & 0.0241 & 1 \\
B & Full & Off gain & FPR & 0.0209 & -0.0420 & 1 \\
B & Full & Off gain & Energy+FPR & 0.0880 & -0.0161 & 2 \\
B & Raw & FG reduction & Energy & 0.1030 & 0.0574 & 1 \\
B & Raw & FG reduction & FPR & 0.1811 & 0.1316 & 1 \\
B & Raw & FG reduction & Energy+FPR & 0.1972 & 0.0825 & 2 \\
B & Full & FG reduction & Energy & 0.1030 & 0.0574 & 1 \\
B & Full & FG reduction & FPR & 0.1687 & 0.1176 & 1 \\
B & Full & FG reduction & Energy+FPR & 0.1794 & 0.0709 & 2 \\
C & Raw & Target gain & Energy & 0.0000 & 0.0000 & 0 \\
C & Raw & Target gain & FPR & 0.0004 & -0.0767 & 1 \\
C & Raw & Target gain & Energy+FPR & 0.0004 & -0.0767 & 1 \\
C & Full & Target gain & Energy & 0.0000 & 0.0000 & 0 \\
C & Full & Target gain & FPR & 0.0019 & -0.0826 & 1 \\
C & Full & Target gain & Energy+FPR & 0.0019 & -0.0826 & 1 \\
C & Raw & Off gain & Energy & 0.0000 & 0.0000 & 0 \\
C & Raw & Off gain & FPR & 0.1063 & -0.0319 & 1 \\
C & Raw & Off gain & Energy+FPR & 0.1063 & -0.0319 & 1 \\
C & Full & Off gain & Energy & 0.0000 & 0.0000 & 0 \\
C & Full & Off gain & FPR & 0.0963 & -0.0426 & 1 \\
C & Full & Off gain & Energy+FPR & 0.0963 & -0.0426 & 1 \\
C & Raw & FG reduction & Energy & 0.0000 & 0.0000 & 0 \\
C & Raw & FG reduction & FPR & 0.2133 & 0.1207 & 1 \\
C & Raw & FG reduction & Energy+FPR & 0.2133 & 0.1207 & 1 \\
C & Full & FG reduction & Energy & 0.0000 & 0.0000 & 0 \\
C & Full & FG reduction & FPR & 0.1956 & 0.1089 & 1 \\
C & Full & FG reduction & Energy+FPR & 0.1956 & 0.1089 & 1 \\
\end{longtable}
\endgroup

\begingroup
\footnotesize
\setlength{\tabcolsep}{4pt}
\begin{longtable}{@{}lllrrrr@{}}
\caption{Incremental PR above Energy. The last two columns are the 95\% bootstrap interval for $\Delta R^2_{\rm LOCO}$, conditional on the saved held-out predictions.}\label{tab:app_fpr_increment}\\
\toprule
Cohort & PR & Outcome & $\Delta R^2$ & $\Delta R^2_{\rm LOCO}$ & CI low & CI high \\
\midrule
\endfirsthead
\multicolumn{7}{l}{\small\itshape Table \thetable\ continued}\\
\toprule
Cohort & PR & Outcome & $\Delta R^2$ & $\Delta R^2_{\rm LOCO}$ & CI low & CI high \\
\midrule
\endhead
\midrule
\multicolumn{7}{r}{\small Continued on next page}\\
\endfoot
\bottomrule
\endlastfoot
A & Raw & Target gain & 0.0884 & 0.0201 & -0.2344 & 0.1290 \\
A & Full & Target gain & 0.1115 & 0.0477 & -0.2156 & 0.1657 \\
A & Raw & Off gain & 0.0480 & 0.0327 & -0.0446 & 0.1432 \\
A & Full & Off gain & 0.0692 & 0.0559 & -0.0387 & 0.1976 \\
A & Raw & FG reduction & 0.0201 & 0.0131 & -0.0289 & 0.0563 \\
A & Full & FG reduction & 0.0268 & 0.0179 & -0.0475 & 0.0656 \\
B & Raw & Target gain & 0.0224 & 0.0155 & -0.0439 & 0.0865 \\
B & Full & Target gain & 0.0106 & -0.0022 & -0.0496 & 0.0423 \\
B & Raw & Off gain & 0.0815 & 0.0092 & -0.2204 & 0.1729 \\
B & Full & Off gain & 0.0279 & -0.0402 & -0.1834 & 0.0563 \\
B & Raw & FG reduction & 0.0942 & 0.0251 & -0.2708 & 0.1389 \\
B & Full & FG reduction & 0.0764 & 0.0135 & -0.2120 & 0.1023 \\
C & Raw & Target gain & 0.0004 & -0.0767 & -0.1649 & -0.0363 \\
C & Full & Target gain & 0.0019 & -0.0826 & -0.1981 & -0.0300 \\
C & Raw & Off gain & 0.1063 & -0.0319 & -0.5470 & 0.1992 \\
C & Full & Off gain & 0.0963 & -0.0426 & -0.5517 & 0.1733 \\
C & Raw & FG reduction & 0.2133 & 0.1207 & -0.6019 & 0.2671 \\
C & Full & FG reduction & 0.1956 & 0.1089 & -0.5220 & 0.2654 \\
\end{longtable}
\endgroup

\begingroup
\footnotesize
\setlength{\tabcolsep}{4pt}
\begin{longtable}{@{}lllrrrl@{}}
\caption{Mean within-checkpoint Spearman correlations for the same fixed cohorts and aggregate outcomes. Defined reports the number of nonconstant checkpoint comparisons.}\label{tab:app_fpr_correlations}\\
\toprule
Cohort & Feature & Outcome & $\rho$ & CI low & CI high & Defined \\
\midrule
\endfirsthead
\multicolumn{7}{l}{\small\itshape Table \thetable\ continued}\\
\toprule
Cohort & Feature & Outcome & $\rho$ & CI low & CI high & Defined \\
\midrule
\endhead
\midrule
\multicolumn{7}{r}{\small Continued on next page}\\
\endfoot
\bottomrule
\endlastfoot
A & Energy & Target gain & -0.3879 & -0.5921 & -0.1573 & 18/18 \\
A & Raw & Target gain & 0.2983 & 0.0477 & 0.5098 & 18/18 \\
A & Full & Target gain & 0.3103 & 0.0727 & 0.5265 & 18/18 \\
A & Energy & Off gain & -0.0082 & -0.2677 & 0.2595 & 18/18 \\
A & Raw & Off gain & -0.0270 & -0.2603 & 0.2243 & 18/18 \\
A & Full & Off gain & -0.0270 & -0.2673 & 0.2103 & 18/18 \\
A & Energy & FG reduction & -0.4769 & -0.6692 & -0.2311 & 15/18 \\
A & Raw & FG reduction & 0.4515 & 0.2085 & 0.6476 & 15/18 \\
A & Full & FG reduction & 0.4522 & 0.2213 & 0.6401 & 15/18 \\
B & Energy & Target gain & -0.0595 & -0.3485 & 0.2587 & 18/18 \\
B & Raw & Target gain & 0.0259 & -0.2967 & 0.3017 & 18/18 \\
B & Full & Target gain & 0.0450 & -0.2474 & 0.3253 & 18/18 \\
B & Energy & Off gain & 0.3238 & 0.0337 & 0.5871 & 18/18 \\
B & Raw & Off gain & -0.2857 & -0.5619 & 0.0159 & 18/18 \\
B & Full & Off gain & -0.2857 & -0.5556 & 0.0033 & 18/18 \\
B & Energy & FG reduction & -0.1298 & -0.4434 & 0.2168 & 11/18 \\
B & Raw & FG reduction & 0.2206 & -0.1340 & 0.5436 & 11/18 \\
B & Full & FG reduction & 0.2219 & -0.1397 & 0.5456 & 11/18 \\
C & Energy & Target gain & NA & NA & NA & 0/18 \\
C & Raw & Target gain & -0.1592 & -0.4926 & 0.1594 & 18/18 \\
C & Full & Target gain & -0.1314 & -0.4573 & 0.1741 & 18/18 \\
C & Energy & Off gain & NA & NA & NA & 0/18 \\
C & Raw & Off gain & 0.0278 & -0.3333 & 0.3889 & 18/18 \\
C & Full & Off gain & 0.0556 & -0.3056 & 0.4167 & 18/18 \\
C & Energy & FG reduction & NA & NA & NA & 0/18 \\
C & Raw & FG reduction & 0.3740 & -0.1111 & 0.7887 & 9/18 \\
C & Full & FG reduction & 0.3740 & -0.1369 & 0.8184 & 9/18 \\
\end{longtable}
\endgroup

\subsection{Correctness transitions and aggregation}
\label{app:transitions}
\label{sec:s5_recovery}
The binary state $ble$ denotes correctness for Base, original LoRA, and an edited adapter, in that order. All eight counts are stored, including zeros.

\begingroup
\small
\setlength{\tabcolsep}{4pt}
\begin{longtable}{@{}lp{.85\linewidth}@{}}
\caption{The eight correctness transitions; categories describe evaluation behavior, not inferred causal mechanisms.}\label{tab:app_transition_definitions}\\
\toprule
State & Interpretation \\
\midrule
\endfirsthead
\multicolumn{2}{l}{\small\itshape Table \thetable\ continued}\\
\toprule
State & Interpretation \\
\midrule
\endhead
\midrule
\multicolumn{2}{r}{\small Continued on next page}\\
\endfoot
\bottomrule
\endlastfoot
000 & All three models are incorrect. \\
001 & New success: both Base and LoRA are incorrect; the edit is correct. \\
010 & A LoRA-only success is lost after editing. \\
011 & A LoRA-only success is retained after editing. \\
100 & A base success lost by LoRA remains incorrect after editing. \\
101 & A base success lost by LoRA is recovered after editing. \\
110 & New damage: Base and LoRA are correct; the edit is incorrect. \\
111 & All three models are correct. \\
\end{longtable}
\endgroup

\begin{equation}
\mathrm{Recovery}=\frac{n_{101}}{n_{100}+n_{101}},\qquad
\mathrm{Retention}=\frac{n_{011}}{n_{010}+n_{011}}.
\label{eq:s5_recovery}
\end{equation}
New-success and new-damage prevalence are $n_{001}/N$ and $n_{110}/N$, respectively, with $N$ the full benchmark sample size. They are not opportunity-conditional rates. Zero opportunity denominators produce NA, not zero. Each aggregation averages defined rates and records coverage: first across the eight commonsense benchmarks, then across the relevant families, then across checkpoints. Overall uses four families, target uses the source family, and off-task uses the remaining three. Conditional Retention is not the Off score.

For a single benchmark, the change in the number of correct answers relative to LoRA is
\begin{equation}
N\,[\mathrm{Acc}(E)-\mathrm{Acc}(L)]
=n_{001}+n_{101}-n_{010}-n_{110}.
\end{equation}
The two conditional rates alone therefore do not determine the accuracy change. Pooled transition counts sum checkpoint--sample events and do not count unique questions; pooled count ratios also differ from the hierarchical macro rates. Scalar-F is included in the performance comparison but not in this fixed three-edit transition analysis.

\paragraph{Recovery with partial preservation.}
Under the overall aggregation, \method{} recovers 62.351\% of base successes lost by LoRA and retains 62.288\% of LoRA-only successes (Table~\ref{tab:app_recovery_roles}). Scalar-E and Flat-E show similar behavior. The \method{}-minus-Scalar-E differences are $-0.524$ Recovery points (95\% paired interval $[-2.587,1.770]$) and $-0.932$ Retention points ($[-2.657,0.969]$). Neither establishes an advantage. These transitions describe recovery with partial preservation across the evaluated edits; they do not identify a mechanism unique to \method{}.

\begingroup
\footnotesize
\setlength{\tabcolsep}{4pt}
\begin{longtable}{@{}llrrrrl@{}}
\caption{Role-specific macro rates (percent) and summed event counts. R/T coverage counts defined checkpoint-level Recovery/Retention rates out of 18; Overall includes target and off-task families.}\label{tab:app_recovery_roles}\label{tab:s5_recovery}\\
\toprule
Role & Method & Recovery & Retention & New success & New damage & R/T \\
\midrule
\endfirsthead
\multicolumn{7}{l}{\small\itshape Table \thetable\ continued}\\
\toprule
Role & Method & Recovery & Retention & New success & New damage & R/T \\
\midrule
\endhead
\midrule
\multicolumn{7}{r}{\small Continued on next page}\\
\endfoot
\bottomrule
\endlastfoot
Overall & HNS & 62.351 & 62.288 & 4553 & 3163 & 18/18 \\
Overall & Scalar-E & 62.876 & 63.220 & 3893 & 3462 & 18/18 \\
Overall & Flat-E & 62.430 & 62.464 & 4487 & 3200 & 18/18 \\
Target & HNS & 58.059 & 72.368 & 306 & 232 & 18/18 \\
Target & Scalar-E & 65.414 & 70.811 & 286 & 265 & 18/18 \\
Target & Flat-E & 59.439 & 72.833 & 308 & 235 & 18/18 \\
Off-task & HNS & 63.782 & 58.928 & 4247 & 2931 & 18/18 \\
Off-task & Scalar-E & 62.030 & 60.689 & 3607 & 3197 & 18/18 \\
Off-task & Flat-E & 63.427 & 59.007 & 4179 & 2965 & 18/18 \\
\end{longtable}
\endgroup

\begingroup
\footnotesize
\setlength{\tabcolsep}{4pt}
\begin{longtable}{@{}lrrrrrrrr@{}}
\caption{All eight pooled overall transition counts. Each method uses the same source checkpoints and evaluation examples.}\label{tab:app_transition_counts}\\
\toprule
Method & 000 & 001 & 010 & 011 & 100 & 101 & 110 & 111 \\
\midrule
\endfirsthead
\multicolumn{9}{l}{\small\itshape Table \thetable\ continued}\\
\toprule
Method & 000 & 001 & 010 & 011 & 100 & 101 & 110 & 111 \\
\midrule
\endhead
\midrule
\multicolumn{9}{r}{\small Continued on next page}\\
\endfoot
\bottomrule
\endlastfoot
HNS & 81930 & 4553 & 10078 & 12267 & 11080 & 16893 & 3163 & 300010 \\
Scalar-E & 82590 & 3893 & 9928 & 12417 & 11883 & 16090 & 3462 & 299711 \\
Flat-E & 81996 & 4487 & 10026 & 12319 & 11105 & 16868 & 3200 & 299973 \\
\end{longtable}
\endgroup

\begingroup
\footnotesize
\setlength{\tabcolsep}{4pt}
\begin{longtable}{@{}llrrrrl@{}}
\caption{Per-benchmark Recovery/Retention macro rates (percent), pooled new-success/damage events, and defined checkpoint coverage. Each benchmark includes all 18 source checkpoints regardless of its target/off-task role.}\label{tab:app_recovery_benchmarks}\\
\toprule
Benchmark & Method & Recovery & Retention & New success & New damage & R/T \\
\midrule
\endfirsthead
\multicolumn{7}{l}{\small\itshape Table \thetable\ continued}\\
\toprule
Benchmark & Method & Recovery & Retention & New success & New damage & R/T \\
\midrule
\endhead
\midrule
\multicolumn{7}{r}{\small Continued on next page}\\
\endfoot
\bottomrule
\endlastfoot
HumanEval & HNS & 57.949 & 68.271 & 70 & 44 & 18/18 \\
HumanEval & Scalar-E & 58.140 & 69.631 & 67 & 40 & 18/18 \\
HumanEval & Flat-E & 58.476 & 67.168 & 71 & 46 & 18/18 \\
GSM8K & HNS & 62.983 & 70.737 & 693 & 487 & 18/18 \\
GSM8K & Scalar-E & 64.978 & 71.324 & 691 & 529 & 18/18 \\
GSM8K & Flat-E & 62.229 & 71.750 & 653 & 478 & 18/18 \\
IFEval & HNS & 67.210 & 55.319 & 226 & 218 & 18/18 \\
IFEval & Scalar-E & 69.129 & 55.211 & 216 & 244 & 18/18 \\
IFEval & Flat-E & 67.794 & 55.685 & 224 & 246 & 18/18 \\
ARC-C & HNS & 60.418 & 68.037 & 204 & 151 & 18/18 \\
ARC-C & Scalar-E & 57.871 & 67.870 & 167 & 182 & 18/18 \\
ARC-C & Flat-E & 60.294 & 68.417 & 208 & 153 & 18/18 \\
ARC-E & HNS & 61.517 & 69.132 & 155 & 123 & 18/18 \\
ARC-E & Scalar-E & 60.243 & 68.692 & 132 & 156 & 18/18 \\
ARC-E & Flat-E & 60.890 & 68.848 & 154 & 129 & 18/18 \\
BoolQ & HNS & 73.364 & 37.012 & 198 & 303 & 18/18 \\
BoolQ & Scalar-E & 71.221 & 40.244 & 194 & 278 & 18/18 \\
BoolQ & Flat-E & 72.484 & 37.741 & 195 & 298 & 18/18 \\
HellaSwag & HNS & 57.839 & 50.269 & 2474 & 1337 & 18/18 \\
HellaSwag & Scalar-E & 54.660 & 50.793 & 1870 & 1522 & 18/18 \\
HellaSwag & Flat-E & 57.512 & 50.247 & 2434 & 1335 & 18/18 \\
OBQA & HNS & 62.114 & 60.503 & 93 & 47 & 18/18 \\
OBQA & Scalar-E & 61.697 & 62.896 & 94 & 42 & 18/18 \\
OBQA & Flat-E & 62.968 & 61.661 & 87 & 43 & 18/18 \\
PIQA & HNS & 64.613 & 50.637 & 101 & 129 & 18/18 \\
PIQA & Scalar-E & 61.042 & 53.555 & 92 & 108 & 18/18 \\
PIQA & Flat-E & 64.680 & 51.095 & 107 & 139 & 18/18 \\
SIQA & HNS & 58.677 & 45.617 & 157 & 187 & 18/18 \\
SIQA & Scalar-E & 58.256 & 48.974 & 140 & 158 & 18/18 \\
SIQA & Flat-E & 58.868 & 45.812 & 167 & 198 & 18/18 \\
WinoGrande & HNS & 51.571 & 57.393 & 182 & 137 & 18/18 \\
WinoGrande & Scalar-E & 49.053 & 60.682 & 230 & 203 & 18/18 \\
WinoGrande & Flat-E & 52.081 & 58.186 & 187 & 135 & 18/18 \\
\end{longtable}
\endgroup

\begingroup
\footnotesize
\setlength{\tabcolsep}{4pt}
\begin{longtable}{@{}lllrrr@{}}
\caption{All method pairs for overall/target/off-task conditional rates. Differences and source-checkpoint 95\% bootstrap intervals are in percentage points; all listed pairs have 18 defined source comparisons.}\label{tab:app_recovery_ci}\\
\toprule
Role & Comparison & Metric & Mean & CI low & CI high \\
\midrule
\endfirsthead
\multicolumn{6}{l}{\small\itshape Table \thetable\ continued}\\
\toprule
Role & Comparison & Metric & Mean & CI low & CI high \\
\midrule
\endhead
\midrule
\multicolumn{6}{r}{\small Continued on next page}\\
\endfoot
\bottomrule
\endlastfoot
Overall & HNS $-$ Scalar-E & Recovery & -0.524 & -2.587 & 1.770 \\
Overall & HNS $-$ Scalar-E & Retention & -0.932 & -2.657 & 0.969 \\
Overall & HNS $-$ Flat-E & Recovery & -0.079 & -0.627 & 0.508 \\
Overall & HNS $-$ Flat-E & Retention & -0.176 & -0.821 & 0.479 \\
Overall & Scalar-E $-$ Flat-E & Recovery & 0.446 & -1.810 & 2.332 \\
Overall & Scalar-E $-$ Flat-E & Retention & 0.756 & -1.063 & 2.292 \\
Target & HNS $-$ Scalar-E & Recovery & -7.355 & -12.235 & -3.038 \\
Target & HNS $-$ Scalar-E & Retention & 1.556 & -1.414 & 4.480 \\
Target & HNS $-$ Flat-E & Recovery & -1.380 & -3.077 & 0.236 \\
Target & HNS $-$ Flat-E & Retention & -0.466 & -1.883 & 0.968 \\
Target & Scalar-E $-$ Flat-E & Recovery & 5.975 & 1.899 & 10.283 \\
Target & Scalar-E $-$ Flat-E & Retention & -2.022 & -4.786 & 0.863 \\
Off-task & HNS $-$ Scalar-E & Recovery & 1.753 & -0.198 & 4.116 \\
Off-task & HNS $-$ Scalar-E & Retention & -1.761 & -3.821 & 0.370 \\
Off-task & HNS $-$ Flat-E & Recovery & 0.355 & -0.453 & 1.106 \\
Off-task & HNS $-$ Flat-E & Retention & -0.079 & -0.714 & 0.635 \\
Off-task & Scalar-E $-$ Flat-E & Recovery & -1.398 & -3.493 & 0.378 \\
Off-task & Scalar-E $-$ Flat-E & Retention & 1.682 & -0.458 & 3.512 \\
\end{longtable}
\endgroup

\subsection{Reproducibility records}
The numerical summaries above use the completed result tables. Preparing these appendix tables involved no retraining or new model inference. The following records separate method construction, evaluation provenance and statistical outputs. They accompany the experiment code; large weights, activations and per-example predictions are not embedded in this appendix. Hash checks establish artifact identity and internal consistency, not independent replication or a causal mechanism.

\begingroup
\footnotesize
\setlength{\tabcolsep}{4pt}
\begin{longtable}{@{}p{.57\linewidth}>{\raggedright\arraybackslash}p{.39\linewidth}@{}}
\caption{Supplementary records under the analysis-section result directory. File names identify machine-readable records. The analysis tables are embedded in this appendix; the separate parameter-control table retains its original external TeX input.}\label{tab:app_file_index}\\
\toprule
Record & Purpose \\
\midrule
\endfirsthead
\multicolumn{2}{l}{\small\itshape Table \thetable\ continued}\\
\toprule
Record & Purpose \\
\midrule
\endhead
\midrule
\multicolumn{2}{r}{\small Continued on next page}\\
\endfoot
\bottomrule
\endlastfoot
\texttt{provenance.json} & Source paths, code provenance, reuse and audit records. \\
\texttt{source\_checkpoints.json} & The 18 retained source identities and configurations. \\
\texttt{evaluation\_sources.tsv} & Evaluation reference/cache lineage and input-identity checks. \\
\texttt{four\_method\_checkpoint\_results.tsv} & Four edited methods, checkpoint scores and eight commonsense components. \\
\texttt{four\_method\_paired\_ci.tsv} & All method-pair performance intervals for overall and fixed groups. \\
\texttt{module\_energy\_audit.tsv} & Per-module analytic/saved energy matching. \\
\texttt{scalar\_f\_module\_audit.tsv} & Per-module Scalar-F matching and energy ratios. \\
\texttt{fpr\_statistics.tsv} & Correlations and Energy/PR regression results for all recorded outcomes. \\
\texttt{fpr\_fold\_membership.tsv} & Source identities included and excluded in each LOCO fold. \\
\texttt{fpr\_leaveout\_audit.json} & Independent check of fold isolation and training-only feature scaling. \\
\texttt{transitions8\_checkpoint\_benchmark.tsv} & All eight correctness-state counts for 594 method--source--benchmark rows. \\
\texttt{transitions8\_summary.tsv} & Hierarchical transition rates, counts and coverage. \\
\texttt{recovery\_paired\_checkpoint\_ci.tsv} & Paired rate intervals across sources on fixed tests. \\
\texttt{recovery\_paired\_sample\_ci.tsv} & Conditional paired test-sample intervals for each checkpoint. \\
\end{longtable}
\endgroup

\section{Additional post-hoc baselines}
\label{app:posthoc_baselines}
\subsection{Construction and evaluation scope}
The extension evaluates three $\epsilon$-PARA configurations and one Spectral Surgery configuration on all 18 original source checkpoints, retaining all three existing training runs. No source adapter is retrained. Current Base, LoRA, and HNS $4+1$ references are used rather than importing scores from the earlier Spectral Surgery paper. Table~\ref{tab:posthoc_baselines} reports every PARA configuration, without selecting one by test-set performance. Paired intervals use 2,000 source-checkpoint bootstrap draws on the fixed test set (Table~\ref{tab:app_posthoc_ci}). These describe paired run heterogeneity, not independent same-recipe training repetitions; the intervals are unadjusted for multiple comparisons.

PARA is implemented through compact QR/SVD; no official implementation was available at the time of this reproduction. Zero-rank modules are excluded from the PEFT configuration, and rank/alpha patterns retain the original scaling. PARA pools singular values across adapter modules and applies a global threshold~\citep{kumaravelu2026para}. With pooled singular values $\{\sigma_{mi}\}$, the retained squared spectral mass satisfies
\begin{equation}
\frac{\sum_{m,i}\sigma_{mi}^2\mathbf{1}[\sigma_{mi}\geq\tau]}
{\sum_{m,i}\sigma_{mi}^2}\geq\epsilon,
\qquad \epsilon\in\{0.90,0.95,0.99\}.
\end{equation}
The remaining gains are unchanged, without subsequent nuclear- or Frobenius-norm restoration. Original LoRA scaling is retained; it is $c=2$ across the source modules in this cohort, so including it would not change the global ordering or energy ratios. Actual retained energy meets each requested budget. Mean parameter retention is 38.16\%, 54.28\%, and 80.16\%, respectively. Parameter retention weights a retained rank in module $m$ by $d_{\mathrm{in},m}+d_{\mathrm{out},m}$, rather than treating ranks in differently sized modules as equal parameter costs.

Spectral Surgery estimates gradient-based sensitivity from 128 training examples with calibration seed 42~\citep{tian2026spectralsurgery}. Calibration provenance, file hashes, and per-checkpoint timing are recorded with the experiment artifacts. It is a calibration-dependent comparator; the main LoRA-Norm edit requires neither these samples nor gradients. The comparison concerns the evaluated complete configurations and is not a controlled ablation of the gradient signal alone. The fixed publication configuration is \texttt{grad\_direction\_}\allowbreak\texttt{residual\_l1\_calib128}, chosen before evaluation from the principal guided-versus-random comparison and matching repository configuration. It uses answer-only teacher-forced loss, mean-absolute normalization, asymmetric multiplicative step sizes $\eta_{\mathrm{suppress}}=2.0$ and $\eta_{\mathrm{enhance}}=0.2$, and nuclear-norm preservation. Editing is restricted to \texttt{o\_proj} and \texttt{down\_proj} in every layer, whereas HNS edits all seven module families; the measured difference therefore cannot be attributed to gradient guidance alone.

\subsection{Execution and numerical audit}
The PARA stage completed 224/224 scheduled cells in 1:43:31 (job 1077); the Spectral Surgery stage completed 80/80 in 0:54:53 (job 1078), starting only after successful completion of the PARA job. These stage counts include their scheduled reference cells and must not be added to the earlier primary count as though all were new unique evaluations. At most two B300 GPUs were used concurrently. Recomputed hashes of all 18 original checkpoint weights and configurations were unchanged. The execution audit reports 12 passing implementation tests, together with successful syntax, compilation, and diff checks.

An initial numerical precheck stopped before evaluation because its FP32 reconstruction error of $1.202\times10^{-5}$ exceeded the initial $10^{-5}$ tolerance. After the discrepancy was attributed to rounding, the tolerance was changed to $5\times10^{-5}$; the measured error was retained in the audit. This threshold change is reported explicitly and does not establish end-to-end prediction equivalence by itself.

Across the Spectral Surgery runs, the recorded calibration-gradient time totals 487.5 seconds and the reported complete editing time totals 1061.2 seconds. The latter is the end-to-end editing time, including gradient calculation, so these totals are not added together. Mean gradient and end-to-end editing times are 27.1 and 59.0 seconds per checkpoint, respectively. No relative editing speedup is claimed without comparable LoRA-Norm timings. Whole-job durations above include evaluation and are not editing-only costs.

\subsection{Interpretation and reproducibility records}
PARA addresses compression as well as prediction, so its parameter savings must accompany the performance comparison. Spectral Surgery's aggregate Target gain with an Off decrease describes this evaluated configuration, rather than a necessary property of all gradient-guided edits. Neither comparison isolates functional energy from spectral shape; the controls in Section~\ref{sec:s5_shape_strength} remain necessary for interpreting that distinction.

The experiment result directory is:
\begin{quote}
\small\texttt{reports/posthoc\_baselines\_para\_spectral\_surgery/}
\end{quote}
Its \texttt{report.md}, \texttt{final\_validation.json}, and
\texttt{scheduler\_jobs.tsv} record the summary, integrity checks, and scheduling.
The method subdirectories \texttt{para/} and \texttt{spectral\_surgery/}
contain \texttt{checkpoint\_results.tsv} and stage reports.

\subsection{Paired comparisons and complete checkpoint results}
\begin{table}[htbp]
\centering\small
\caption{LoRA-Norm minus each new baseline: cohort means and 95\% paired source-checkpoint bootstrap intervals, in percentage points. Positive differences favor LoRA-Norm for Target/Off; negative differences favor it for FG. Intervals are descriptive and unadjusted.}
\label{tab:app_posthoc_ci}
\begin{tabular}{llrrr}
\toprule
Comparator & Metric & Difference & CI low & CI high\\
\midrule
PARA $\epsilon=.90$ & Target & +3.306 & +1.496 & +5.348 \\
PARA $\epsilon=.90$ & Off & +2.779 & +1.506 & +4.116 \\
PARA $\epsilon=.90$ & FG & -2.027 & -3.129 & -1.080 \\
PARA $\epsilon=.95$ & Target & +3.501 & +2.114 & +5.063 \\
PARA $\epsilon=.95$ & Off & +3.156 & +1.757 & +4.547 \\
PARA $\epsilon=.95$ & FG & -2.351 & -3.502 & -1.401 \\
PARA $\epsilon=.99$ & Target & +3.641 & +2.415 & +5.039 \\
PARA $\epsilon=.99$ & Off & +3.014 & +1.705 & +4.357 \\
PARA $\epsilon=.99$ & FG & -2.220 & -3.383 & -1.313 \\
Spectral Surgery & Target & +3.320 & +2.113 & +4.500 \\
Spectral Surgery & Off & +4.510 & +3.115 & +5.784 \\
Spectral Surgery & FG & -3.378 & -4.486 & -2.225 \\
\bottomrule
\end{tabular}
\end{table}

\begingroup
\small
\setlength{\tabcolsep}{4pt}
\begin{longtable}{lll lrrr}
\caption{Complete new-baseline results for all 18 source checkpoints. Target and Off are percentages; FG is in percentage points. Code, Math, and IF denote Magicoder, MetaMath, and Tulu training, respectively.}\label{tab:app_posthoc_full}\\
\toprule
Base & Task & Run & Method & Target & Off & FG\\
\midrule
\endfirsthead
\toprule
Base & Task & Run & Method & Target & Off & FG\\
\midrule
\endhead
\bottomrule
\endfoot
Llama & Code & 42 & PARA $\epsilon=.90$ & 56.707 & 61.146 & 4.152 \\
Llama & Code & 42 & PARA $\epsilon=.95$ & 54.878 & 61.489 & 3.809 \\
Llama & Code & 42 & PARA $\epsilon=.99$ & 53.659 & 60.559 & 4.740 \\
Llama & Code & 42 & Spectral Surgery & 54.268 & 59.375 & 5.924 \\
Llama & Code & 43 & PARA $\epsilon=.90$ & 56.098 & 61.743 & 4.439 \\
Llama & Code & 43 & PARA $\epsilon=.95$ & 54.878 & 61.690 & 4.315 \\
Llama & Code & 43 & PARA $\epsilon=.99$ & 54.878 & 61.070 & 4.683 \\
Llama & Code & 43 & Spectral Surgery & 56.098 & 58.750 & 6.548 \\
Llama & Code & 44 & PARA $\epsilon=.90$ & 57.317 & 62.439 & 4.067 \\
Llama & Code & 44 & PARA $\epsilon=.95$ & 54.878 & 61.853 & 4.436 \\
Llama & Code & 44 & PARA $\epsilon=.99$ & 54.268 & 62.323 & 4.441 \\
Llama & Code & 44 & Spectral Surgery & 54.878 & 59.666 & 5.632 \\
Llama & Math & 42 & PARA $\epsilon=.90$ & 76.422 & 59.265 & 3.933 \\
Llama & Math & 42 & PARA $\epsilon=.95$ & 76.573 & 58.545 & 4.247 \\
Llama & Math & 42 & PARA $\epsilon=.99$ & 76.952 & 58.052 & 4.537 \\
Llama & Math & 42 & Spectral Surgery & 76.573 & 58.156 & 4.636 \\
Llama & Math & 43 & PARA $\epsilon=.90$ & 75.512 & 58.665 & 3.924 \\
Llama & Math & 43 & PARA $\epsilon=.95$ & 75.436 & 58.711 & 3.878 \\
Llama & Math & 43 & PARA $\epsilon=.99$ & 74.905 & 58.287 & 4.099 \\
Llama & Math & 43 & Spectral Surgery & 74.526 & 57.920 & 4.669 \\
Llama & Math & 44 & PARA $\epsilon=.90$ & 75.588 & 59.550 & 3.243 \\
Llama & Math & 44 & PARA $\epsilon=.95$ & 75.663 & 59.542 & 3.657 \\
Llama & Math & 44 & PARA $\epsilon=.99$ & 75.133 & 59.555 & 3.643 \\
Llama & Math & 44 & Spectral Surgery & 74.223 & 59.301 & 3.694 \\
Llama & IF & 42 & PARA $\epsilon=.90$ & 63.216 & 65.891 & 0.000 \\
Llama & IF & 42 & PARA $\epsilon=.95$ & 64.880 & 62.450 & 3.284 \\
Llama & IF & 42 & PARA $\epsilon=.99$ & 63.031 & 66.119 & 0.044 \\
Llama & IF & 42 & Spectral Surgery & 63.031 & 58.662 & 6.260 \\
Llama & IF & 43 & PARA $\epsilon=.90$ & 62.847 & 67.342 & 0.000 \\
Llama & IF & 43 & PARA $\epsilon=.95$ & 62.477 & 66.673 & 0.000 \\
Llama & IF & 43 & PARA $\epsilon=.99$ & 64.510 & 67.289 & 0.000 \\
Llama & IF & 43 & Spectral Surgery & 64.325 & 65.851 & 0.682 \\
Llama & IF & 44 & PARA $\epsilon=.90$ & 62.477 & 67.692 & 0.000 \\
Llama & IF & 44 & PARA $\epsilon=.95$ & 62.662 & 67.916 & 0.000 \\
Llama & IF & 44 & PARA $\epsilon=.99$ & 63.771 & 66.782 & 0.303 \\
Llama & IF & 44 & Spectral Surgery & 63.031 & 58.315 & 7.961 \\
Qwen & Code & 42 & PARA $\epsilon=.90$ & 64.024 & 78.034 & 1.701 \\
Qwen & Code & 42 & PARA $\epsilon=.95$ & 66.463 & 78.158 & 1.556 \\
Qwen & Code & 42 & PARA $\epsilon=.99$ & 65.854 & 77.517 & 2.223 \\
Qwen & Code & 42 & Spectral Surgery & 68.902 & 75.713 & 3.937 \\
Qwen & Code & 43 & PARA $\epsilon=.90$ & 60.366 & 77.862 & 1.856 \\
Qwen & Code & 43 & PARA $\epsilon=.95$ & 62.195 & 76.833 & 2.893 \\
Qwen & Code & 43 & PARA $\epsilon=.99$ & 64.024 & 77.480 & 2.229 \\
Qwen & Code & 43 & Spectral Surgery & 65.244 & 75.042 & 4.354 \\
Qwen & Code & 44 & PARA $\epsilon=.90$ & 65.854 & 78.148 & 1.540 \\
Qwen & Code & 44 & PARA $\epsilon=.95$ & 65.854 & 77.886 & 1.783 \\
Qwen & Code & 44 & PARA $\epsilon=.99$ & 65.854 & 77.987 & 1.685 \\
Qwen & Code & 44 & Spectral Surgery & 67.073 & 75.418 & 3.951 \\
Qwen & Math & 42 & PARA $\epsilon=.90$ & 84.913 & 72.896 & 0.610 \\
Qwen & Math & 42 & PARA $\epsilon=.95$ & 84.534 & 72.151 & 0.899 \\
Qwen & Math & 42 & PARA $\epsilon=.99$ & 84.079 & 71.848 & 1.269 \\
Qwen & Math & 42 & Spectral Surgery & 84.306 & 73.076 & 0.308 \\
Qwen & Math & 43 & PARA $\epsilon=.90$ & 84.230 & 69.133 & 4.071 \\
Qwen & Math & 43 & PARA $\epsilon=.95$ & 83.245 & 69.093 & 4.139 \\
Qwen & Math & 43 & PARA $\epsilon=.99$ & 83.851 & 70.384 & 2.809 \\
Qwen & Math & 43 & Spectral Surgery & 83.927 & 72.156 & 1.103 \\
Qwen & Math & 44 & PARA $\epsilon=.90$ & 83.927 & 64.005 & 9.134 \\
Qwen & Math & 44 & PARA $\epsilon=.95$ & 83.397 & 63.458 & 9.621 \\
Qwen & Math & 44 & PARA $\epsilon=.99$ & 84.230 & 63.638 & 9.442 \\
Qwen & Math & 44 & Spectral Surgery & 84.382 & 65.739 & 7.348 \\
Qwen & IF & 42 & PARA $\epsilon=.90$ & 70.795 & 86.125 & 0.000 \\
Qwen & IF & 42 & PARA $\epsilon=.95$ & 69.131 & 85.674 & 0.000 \\
Qwen & IF & 42 & PARA $\epsilon=.99$ & 67.652 & 86.132 & 0.000 \\
Qwen & IF & 42 & Spectral Surgery & 68.022 & 85.568 & 0.000 \\
Qwen & IF & 43 & PARA $\epsilon=.90$ & 70.055 & 85.450 & 0.000 \\
Qwen & IF & 43 & PARA $\epsilon=.95$ & 69.501 & 85.034 & 0.000 \\
Qwen & IF & 43 & PARA $\epsilon=.99$ & 67.098 & 85.407 & 0.000 \\
Qwen & IF & 43 & Spectral Surgery & 67.837 & 85.203 & 0.000 \\
Qwen & IF & 44 & PARA $\epsilon=.90$ & 67.652 & 84.840 & 0.000 \\
Qwen & IF & 44 & PARA $\epsilon=.95$ & 67.837 & 86.286 & 0.000 \\
Qwen & IF & 44 & PARA $\epsilon=.99$ & 68.207 & 85.566 & 0.000 \\
Qwen & IF & 44 & Spectral Surgery & 67.098 & 85.164 & 0.000 \\
\end{longtable}
\endgroup

\clearpage
\section{Sensitivity to HNS iteration counts}
\label{app:hns_step_grid}

\subsection{Protocol and complete grid}
To match the primary adaptation-task scope, we report six source LoRA adapters from the archived 2026-09-12 sweep: Qwen3-8B and Llama-3.1-8B-Instruct, each adapted on Magicoder, MetaMath, and Tulu. There is one source checkpoint per backbone--training-task pair. The original eight-checkpoint archive is retained with the experiment artifacts; its two Commonsense170K-trained sources are excluded from all tables and aggregates in this section. The nine nonzero settings are $(K_f,K_s)\in\{2,4,8\}\times\{0,1,2\}$ in Eq.~\eqref{eq:hns_iteration}. A label such as $4+1$ denotes four fast steps followed by one stable step, not an adapter rank. Every nonzero setting edits all LoRA modules at strength 1, retains rank 16 and the original adapter scaling, and restores each module's nuclear norm. Both $4+1$ and $8+2$ were rebuilt and evaluated with the other variants.

Only target-task performance is evaluated: HumanEval (164 problems), GSM8K (1,319 problems), and IFEval (541 prompts), respectively. Greedy inference uses generation seed 42 and batch-invariant kernels, with maximum generated lengths 512/512/2,048. The retained run used 65,536 maximum batched tokens. These are six archived sources with one evaluation per setting, not three training repetitions; generation seed 42 does not establish the sources' training seeds. This block retains its own Base, LoRA, and $0+0$ reconstruction references and is not pooled with the current 18-checkpoint evaluation. It provides neither an Off/FG parameter sweep nor a rank or editing-strength ablation.

\begin{table}[!htbp]
\centering
\small
\setlength{\tabcolsep}{4pt}
\caption{Complete step grid for the six archived checkpoints matching the primary adaptation tasks (percentages). HE/GSM/IF denote HumanEval, GSM8K, and IFEval, evaluated on their corresponding training-task adapters. The bold row label identifies the main $4+1$ setting, not a score maximum.}
\label{tab:hns_step_grid_full}
\begin{tabular}{@{}lrrrrrr@{}}
\toprule
& \multicolumn{3}{c}{Qwen3-8B} & \multicolumn{3}{c}{Llama-3.1-8B-Instruct}\\
\cmidrule(lr){2-4}\cmidrule(lr){5-7}
Setting & HE & GSM & IF & HE & GSM & IF\\
\midrule
Base & 66.46 & 85.67 & 70.61 & 52.44 & 62.40 & 62.11\\
LoRA & 66.46 & 84.15 & 67.65 & 53.66 & 77.10 & 63.22\\
$0+0$ (rebuild) & 67.68 & 84.53 & 67.65 & 54.27 & 77.33 & 63.59\\
\midrule
$2+0$ & 75.61 & 88.25 & 70.06 & 53.66 & 80.89 & 65.06\\
$2+1$ & 74.39 & 88.25 & 70.79 & 54.27 & 80.82 & 65.06\\
$2+2$ & 76.22 & 88.25 & 70.06 & 53.66 & 80.29 & 64.51\\
$4+0$ & 74.39 & 88.25 & 71.35 & 53.66 & 80.89 & 65.43\\
\textbf{4+1} & 75.00 & 88.02 & 70.24 & 54.88 & 80.67 & 64.88\\
$4+2$ & 75.61 & 88.17 & 70.43 & 53.66 & 81.05 & 65.06\\
$8+0$ & 75.00 & 88.40 & 70.61 & 54.27 & 80.67 & 64.70\\
$8+1$ & 75.00 & 88.55 & 70.43 & 54.88 & 80.14 & 64.33\\
$8+2$ & 75.61 & 88.48 & 68.95 & 53.66 & 80.82 & 65.25\\
\bottomrule
\end{tabular}
\end{table}

\paragraph{Reconstruction control.}
The $0+0$ variant performs compact SVD and balanced reconstruction without a nontrivial spectral edit. It is an identity at the update level in exact arithmetic, but the stored factors and numerical execution differ. Its HumanEval changes are $+1.22$ points for Qwen and $+0.61$ for Llama (two and one problems), while changes on the other four checkpoints range from $0$ to $+0.3791$ points. Across all six checkpoints, its mean gain is $+0.47$ points, with five wins, one tie, and no losses. We retain this control explicitly rather than treating reconstructed factors as numerically interchangeable with the original LoRA.

\paragraph{Task-dependent response.}
Both GSM8K checkpoints improve under every nonzero setting. Each nonzero setting improves five or six of the selected checkpoints and decreases none; the remaining cases are ties on Llama HumanEval. The main $4+1$ setting improves all six checkpoints. The descriptive maxima differ across tasks: Qwen HumanEval favors $2+2$, Qwen GSM8K favors $8+1$, Llama GSM8K favors $4+2$, and both IFEval checkpoints favor $4+0$; Llama HumanEval ties at $4+1$ and $8+1$. These maxima are observations on the evaluated test sets, not independently validated configuration choices.

\clearpage
\subsection{Fixed-configuration aggregates and uncertainty}
Table~\ref{tab:hns_step_aggregate} averages each fixed setting equally over the same six checkpoints; it never selects a different setting for each checkpoint. The observed mean improvement is $+3.42$--$+3.62$ points across the nonzero grid, and the main $4+1$ setting yields $+3.57$ points. Settings $4+0$ and $4+2$ both round to $+3.62$ points; the unrounded maximum is $4+2$, only $0.00024$ points above $4+0$ and $0.04844$ points above $4+1$. These small observed differences do not establish significant differences between settings. Increasing either iteration count does not yield monotonic task performance. We use this sweep to describe sensitivity while retaining $4+1$ as the common configuration in the main experiments.

\begin{table}[!htbp]
\centering
\small
\setlength{\tabcolsep}{10pt}
\caption{Fixed-setting gains over original LoRA on the six archived checkpoints matching the primary adaptation tasks (percentage points). Means and medians weight checkpoints equally; wins/ties/losses use unrounded scores.}
\label{tab:hns_step_aggregate}
\begin{tabular}{@{}lrrc@{}}
\toprule
Setting & Mean gain & Median gain & Win / tie / loss\\
\midrule
$0+0$ (rebuild) & +0.47 & +0.37 & 5 / 1 / 0\\
\midrule
$2+0$ & +3.55 & +3.10 & 5 / 1 / 0\\
$2+1$ & +3.56 & +3.43 & 6 / 0 / 0\\
$2+2$ & +3.46 & +2.79 & 5 / 1 / 0\\
$4+0$ & +3.62 & +3.74 & 5 / 1 / 0\\
\textbf{4+1} & +3.57 & +3.08 & 6 / 0 / 0\\
$4+2$ & +3.62 & +3.36 & 5 / 1 / 0\\
$8+0$ & +3.57 & +3.26 & 6 / 0 / 0\\
$8+1$ & +3.51 & +2.90 & 6 / 0 / 0\\
$8+2$ & +3.42 & +2.87 & 5 / 1 / 0\\
\bottomrule
\end{tabular}
\end{table}

\paragraph{Paired inference.}
The archived analysis uses 20,000 paired bootstrap draws of example-level correctness differences. These percentile intervals describe test-example uncertainty for fixed source adapters; they are not training-seed intervals. Exact two-sided McNemar tests are adjusted by Holm's method over ten comparisons against original LoRA within each checkpoint, including $0+0$. Restricting the reported cohort leaves these within-checkpoint comparisons unchanged, so the retained intervals and adjusted $p$-values are reused without recomputation.

\begin{table}[!htbp]
\centering
\small
\setlength{\tabcolsep}{7pt}
\caption{Archived $4+1$ versus original LoRA on the six primary-task checkpoints: gain and ordinary paired 95\% interval in percentage points. $p_{\mathrm{Holm}}$ adjusts ten within-checkpoint comparisons. Per-checkpoint intervals and adjusted $p$-values are unchanged by the cohort restriction.}
\label{tab:hns_step_anchor_stats}
\begin{tabular}{@{}llrrr@{}}
\toprule
Base & Target & Gain & 95\% interval & $p_{\mathrm{Holm}}$\\
\midrule
Qwen & HumanEval & +8.54 & $[1.22, 15.85]$ & 0.213\\
Qwen & GSM8K & +3.87 & $[2.12, 5.69]$ & $<0.001$\\
Qwen & IFEval & +2.59 & $[0.00, 5.18]$ & 0.448\\
\midrule
Llama & HumanEval & +1.22 & $[-4.27, 6.71]$ & 1.000\\
Llama & GSM8K & +3.56 & $[1.74, 5.38]$ & $<0.001$\\
Llama & IFEval & +1.66 & $[-1.48, 4.81]$ & 1.000\\
\bottomrule
\end{tabular}
\end{table}

Across all nine nonzero settings, both GSM8K checkpoints have positive paired intervals and Holm-adjusted $p<0.05$. Qwen HumanEval also has positive ordinary intervals, but none of its comparisons survives the ten-way correction. All Llama HumanEval and Llama IFEval intervals include zero. For Qwen IFEval, ordinary intervals exclude zero at $2+1$, $4+0$, and $8+1$, but no comparison survives the ten-way correction. Table~\ref{tab:hns_step_anchor_stats} illustrates the fixed $4+1$ setting. All tests compare a setting with original LoRA, so they do not establish significant differences or equivalence between HNS settings. The full score and statistical tables are retained with the experiment artifacts.
\FloatBarrier

\clearpage
\section{Exploratory scalar calibration and numerical recheck}
\label{app:scalar_calibration}

\paragraph{Scope and representation.}
This archived recheck uses one Qwen3-8B--MetaMath source and an all-module HNS reference, specified as $8+2$ in the archive's method summary rather than the main $4+1$. Reconstructed variants share a float32 source SVD and single-sided factors:
\[
A_m'=V_m^\top,\qquad B_m'=U_m\operatorname{diag}(d_m).
\]
Only $d_m$ changes: the source spectrum for zero rebuild, a scalar multiple for scaling controls, or the source-aligned archived HNS gains. All 252 modules retain original LoRA scaling. This representation differs from balanced reconstruction; these results are not pooled with the primary 18-source cohort. Existing-representation HNS is diagnostic only.

\paragraph{Coefficient selection.}
A fixed 256-question calibration subset selected a global multiplier from Table~\ref{tab:app_scalar_grid} by strict accuracy, with ties favoring the value closest to one. The selected coefficient was evaluated without reselection on a separate 512-question subset. Strict extracted-answer equality is primary; numeric equivalence is secondary.

\begin{table}[!htbp]
\centering
\small
\caption{Archived global-scalar calibration on 256 questions. Scores are percentages.}
\label{tab:app_scalar_grid}
\begin{tabular}{@{}rrr@{}}
\toprule
$\gamma$ & Strict & Numeric \\
\midrule
1.00 & 84.77 & 85.16 \\
0.85 & 85.55 & 86.72 \\
0.70 & 85.94 & 87.50 \\
0.60 & 85.16 & 86.72 \\
0.50 & 86.72 & 88.67 \\
0.40 & 88.67 & 91.02 \\
\bottomrule
\end{tabular}
\end{table}

The selected $\gamma=0.40$ is the lower boundary of the fixed grid, so this experiment does not identify the scalar optimum. It is also distinct from matching the archived HNS adapter's total Frobenius norm, which gives $\gamma\approx0.609$. The per-module control instead scales each source spectrum by $\gamma_m=\|t_m^{\mathrm{archived\ HNS}}\|_2/\|\sigma_m\|_2$, matching that module's archived HNS Frobenius norm without matching functional energy.

\begin{table}[!htbp]
\centering
\small
\caption{Archived 512-question recheck. Common-basis variants share their source singular vectors and factorization path. Scores are percentages.}
\label{tab:app_scalar_recheck}
\begin{tabular}{@{}lrr@{}}
\toprule
Variant & Strict & Numeric \\
\midrule
Original LoRA & 83.40 & 84.57 \\
Common-basis zero rebuild & 82.81 & 83.98 \\
Common-basis global scalar ($\gamma=0.40$) & 85.74 & 88.87 \\
Common-basis per-module norm-matched scalar & 84.77 & 86.72 \\
Common-basis archived HNS & 86.13 & 88.67 \\
Archived HNS in its existing representation & 86.72 & 89.45 \\
\bottomrule
\end{tabular}
\end{table}

Common-basis HNS minus selected global scaling is $+0.39$ strict points (paired question-bootstrap 95\% interval $[-1.56,2.34]$; exact McNemar $p=0.839$) and $-0.20$ numeric points ($[-1.76,1.56]$). Against the common-basis per-module scalar, the strict difference is $+1.37$ points ($[-0.78,3.52]$). These comparisons do not establish HNS superiority or statistical equivalence.

\paragraph{Numerical and evidential limits.}
Identical adapters produced identical sequences across three fresh processes on a 32-question diagnostic enriched with previously changed outputs. Alternative factorizations nevertheless changed some sequences. Zero rebuild shifts strict accuracy by $-0.59$ points on the 512-question subset: repeatability does not establish representation invariance, and this shift is not a universal noise floor.

Earlier evaluations already used all 1,319 GSM8K test questions. This is an exploratory one-source target-task recheck, not independent downstream confirmation or a clean held-out selection study. It does not establish retention benefits or superiority over independently validated scaling.
\FloatBarrier

\end{document}